\pdfoutput=1

\documentclass[11pt]{article}

\usepackage[final]{acl}

\usepackage{times}
\usepackage{latexsym}
\usepackage{mathtools}
\usepackage{float}

\usepackage[T1]{fontenc}

\usepackage[utf8]{inputenc}

\usepackage{microtype}

\usepackage{inconsolata}

\usepackage{graphicx}
\usepackage{hyperref}

\title{Efficient Aspect-Based Summarization of Climate Change Reports with Small Language Models}

\author{Iacopo Ghinassi \\
  \small{Queen Mary University of London / London, UK} \\
  \small{\texttt{i.ghinassi@qmul.ac.uk}} \\\And
  Leonardo Catalano \\
  \small{Independent Researcher}\\
  \small{\texttt{leonardocatalano995@gmail.com}} \\
  \\\And
  Tommaso Colella \\
  \small{University of Pisa / Pisa, Italy}\\
  \small{\texttt{t.colella@studenti.unipi.it}} \\}

\begin{document}
\maketitle
\begin{abstract}
The use of Natural Language Processing (NLP) for helping decision-makers with Climate Change action has recently been highlighted as a use case aligning with a broader drive towards NLP technologies for social good. In this context, Aspect-Based Summarization (ABS) systems that extract and summarize relevant information are particularly useful as they provide stakeholders with a convenient way of finding relevant information in expert-curated reports. In this work, we release a new dataset for ABS of Climate Change reports and we employ different Large Language Models (LLMs) and so-called Small Language Models (SLMs) to tackle this problem in an unsupervised way. Considering the problem at hand, we also show how SLMs are not significantly worse for the problem while leading to reduced carbon footprint; we do so by applying for the first time an existing framework considering both energy efficiency and task performance to the evaluation of zero-shot generative models for ABS. Overall, our results show that modern language models, both big and small, can effectively tackle ABS for Climate Change reports but more research is needed when we frame the problem as a Retrieval Augmented Generation (RAG) problem and our work and dataset will help foster efforts in this direction.\footnote{Find dataset at: \url{https://huggingface.co/datasets/ighina/SumIPCC} And code at: \url{https://github.com/Ighina/LLMClimate2024}}
\end{abstract}

\section{Introduction}

Climate change reports are critical for policy-makers and researchers in tackling climatic challenges and, as such, summarization of such reports is a task in line with recent work advocating for ways in which Natural Language Processing (NLP) can help climate scientists and policy-makers and make a positive impact \cite{stede-patz-2021-climate}.

When dealing with such information-dense documents, however, simple summarization might be too limiting, as the end user might need a summary with respect to a specific topic. Summarizing a text with respect to a specific aspect or topic is known as Aspect-Based Summarization (ABS) and it has a long history in NLP \cite{titov-mcdonald-2008-joint}. 

Recently, the landscape of NLP has seen a revolution happening in the form of Large Language Models (LLMs), which are capable of performing many tasks without training, therefore being particularly useful in under-resourced domains like the one of climate change reports \cite{ziyu-etal-2023-lens}. 
These models, however, comprise billions of parameters, and, as such, their carbon footprint is one of the main factors leading to criticisms of their use \cite{2024ICLR:LLMCarbon}, while relatively smaller LLMs, labeled as Small Language Models (SLMs), have started gaining traction in the literature \cite{ranaldi-freitas-2024-aligning}.
In this work, then, we show that LLMs and SLMs can be successfully applied to the task of ABS in the context of Climate Change reports.


The main questions informing our work are:\par
\textbf{Q1}: Can LLMs and SLMs successfully perform ABS of Climate Change reports and how do they compare to other unsupervised alternatives?\par
\textbf{Q2}: are SLMs comparable in performance to larger LLMs for our task?\par
\textbf{Q3}: how do our models' performance deteriorate in the absence of ground truth paragraphs to summarize?

Our main contributions then are:\par
1) We evaluate LLMs and SLMs in the context of ABS for Climate Change reports with ground truth paragraphs and within a RAG setting, and we introduce a new dataset for this task.\par
2) We focus on energy efficiency and we adapt an existing framework for energy-aware summarization evaluation to LLMs and SLMs for the first time, showing how the latter can perform similarly to the first for the task at hand and lead to massive energy saving.

\section{Related Work}
\subsection{NLP and Climate Change}
NLP can help with a variety of problems related to Climate Change including but not limited to: climate stance detection \cite{fraile-hernandez-penas-2024-hamison}, climate-related question answering \cite{Vaghefi2023-fc, nlp4pi-2022-nlp} and automatic fact-checking \cite{su141811724, fakenews-det}. NLP can also improve access to information, which can be used for educational or policy-making purposes \cite{stede-patz-2021-climate}.

Our contribution, then, points in this direction and it builds on previous work to assess a new task in the area, namely that of ABS. Previous work, in fact, has drawn from data similar to the one we use in order to create a chatbot that can answer questions related to climate change with access to the most up-to-date information \cite{Vaghefi2023-fc}. As new reports and new knowledge get produced at a fast pace, however, the need to assess the zero-shot ability of LLMs to summarize such reports in an efficient and fine-grained way is crucial to further help their reading from both policy-makers and researchers. No research in this direction exists in our knowledge and our work aims to fill this gap.

\subsection{Aspect-Based Summarization}
ABS is the task of summarizing a given text with respect to a specific aspect or topic \cite{titov-mcdonald-2008-joint}. The task is particularly useful in aiding the reading of complex, multi-topic content such as news bulletins \cite{frermann-klementiev-2019-inducing} or Wikipedia articles \cite{10.1162/tacl_a_00362}. 


In the context of ABS, the models developed for the task falls broadly in the category of supervised \cite{tan-etal-2020-summarizing, supervised-aspect-based, ahuja-etal-2022-aspectnews} and unsupervised models \cite{soleimani-etal-2022-zero, coavoux-etal-2019-unsupervised}, where the firsts have shown improvements over the latter, but do need a sufficient number of training samples, for which there is a scarcity of data, especially in certain domains \cite{yang-etal-2023-oasum}. More recently, modern LLMs have shown performance on par with previous supervised models also in unsupervised (i.e. zero-shot) setting for various NLP tasks \cite{ziyu-etal-2023-lens} including summarization \cite{10.1162/tacl_a_00632}. Such models are mostly under-explored in the context of ABS, as just isolated examples of their use for the task exist in the literature, which does not present comparisons between LLMs and SLMs and is limited to hotel reviews summarization \cite{su16041640, bhaskar-etal-2023-prompted}.

\subsection{SLMs and Efficiency Evaluation}
Modern LLMs are extremely effective for a variety of tasks, but they comprise billions of parameters, leading to consideration of efficiency and environmental externalities associated with their use \cite{Tokayev_2023}. These concerns have led to consider the overall environmental cost of such models when deploying them \cite{2024ICLR:LLMCarbon}.

At the same time, in the last year much effort has been spent in making the LLM landscape more efficient \cite{wan2024efficient}, either by proposing SLMs, yielding comparable results to LLMs thanks to refined datasets and knowledge distillation \cite{abdin2024phi3, gemmateam2024gemma, gu2024minillm}, or by exploring different types of quantization which can diminish the computational burden while maintaining a good trade-off with performance \cite{Yao_Wu_Li_Youn_He_2024} or both.

Recent literature has proposed to include models' efficiency in evaluating summarization \cite{Moro_Ragazzi_Valgimigli_2023}, but without including LLMs in their experiments. Much NLP literature has often ignored considerations about model efficiency, but as the models get bigger and the marginal improvements get smaller, including model efficiency in the evaluation is important for more sustainable and, ultimately, more usable NLP systems.

In this work, then, we draw also on literature on SLMs and efficiency evaluation in developing our experiments and then assessing them.

\section{Methodology}
\subsection{Zero-Shot Aspect-Based Summarization with LLMs}
In order to perform ABS with out-of-the-box LLMs and SLMs, we developed a simple prompt template which is presented to each model for a fair comparison. The prompt template $T$ has the following format:

$T$="Summarize the main takeaways from the \\following text with respect to topic \{topic\}. Text: \{text\}"

We define the substitution function $sub$, which takes as inputs the template $T$, $topic$ and $text$ and substitutes \{topic\} and \{text\} in $T$ with $topic$ and $text$, respectively, thus obtaining:
\begin{equation}
    prompt=sub(T, topic, text)
    \label{eq:prompt}
\end{equation}

As we will see below, at times more than one paragraph needs to be summarized. Defining the collection of paragraphs to be summarized $P=\{p_1, ..., p_n\}$, where $p_i$ are the individual paragraphs, we obtain:
\begin{equation}
    text=\begin{cases}
    P, |P|=1\\
    concat(P), |P|>1
\end{cases}
\label{eq:cases}
\end{equation}

where $concat$ indicates the concatenation of all the paragraphs in $P$.

The generation process, then, is done as:
\begin{equation}
    \hat{y}=LLM(prompt)
    \label{eq:gen}
\end{equation}
Where $LLM$ is the LLM currently used and $\hat{y}$ is the generated summary.

In many cases, there is also a limitation in the number of maximum tokens that some of the models can accept and especially in the case of many paragraphs $p$ to be summarized the length of the input text might exceed this limit. We have tackled these instances by applying an iterative procedure where we summarize individual paragraphs and then we ask the given LLM to summarize the concatenation of the summaries.




We formally define this procedure in Appendix \ref{sec:appendixC}, together with the implications on the performance of such cases.

\subsection{Retrieval Augmented Generation}
To answer Q3 and test the limits of our approach, we also investigate Retrieval Augmented Generation (RAG), where we automatically retrieve the $k$ most relevant paragraphs from the given climate report and we use them as input for the LLM, instead of the ground truth paragraphs. This setting relates to the real-world use case in which, e.g., a policymaker wants an automatic system to both find the relevant information in the report and summarize it. Formally, we define an encoder model $enc$ such that it encodes all the reports' paragraphs $p_i$ as:
\begin{equation}
    e_i=enc(p_i), e_i\in\Re^d
\end{equation}
with $d$ being the dimensionality of the embeddings from the given encoder $enc$. At inference time, the given aspect or topic $topic$ is encoded in the same embeddings space as:

\begin{equation}
    q=enc(topic), q\in\Re^d
    \label{eq:query_emb}
\end{equation}

At this point, we define a number $k$ of paragraphs that we want to retrieve from the collection of all paragraph indices $P_{ind}=\{1, ..., N\}$ and we retrieve the subset of paragraph indices $P_{sub}\subset P$ as:
\begin{equation}
    P_{sub}=argmax_{i\in P_{ind}}(cos(q, e_i)), s.t. |P_{sub}|=k
\end{equation}

where $cos$ represents the cosine distance between the query embedding $q$ and the given paragraph embedding $e_i$.

Having obtained the paragraphs associated with their indices in $P_{sub}$, we then obtain $text$ as described in equation \ref{eq:cases}. The final summary $\hat{y}$ is then obtained as:

\begin{equation}
    \hat{y}=LLM(prompt_{rag})
\end{equation}

where $prompt_{rag}$ is obtained either with equation \ref{eq:prompt} or with equations \ref{eq:long} and \ref{eq:long2} according to whether $text$ is longer than the character threshold as explained above.



\subsection{Extractive Summarization Baseline}
To compare the performance of LLMs with a non-generative baseline, we develop a simple extractive approach, based on the understanding of the task as a question-answering task. For each example, we again define an encoder $enc$ and we follow equation \ref{eq:query_emb} to obtain a query embedding $q$. Having obtained $text$ in one of the ways previously defined, we then divide it into sentences with the method by \citet{punkt} and group them as $S=\{s_1,...,s_n\}$ with $n$ being the number of sentences in $text$. Each sentence $s_i$ is then encoded as:

\begin{equation}
    e_s^i=enc(s_i), e_s^i\in \Re^d
\end{equation}

We define a number $k$ of sentences to be extracted and the collection of all sentence indices in the document $S_{ind}=\{1,...,n\}$ and we obtain its subset $S_{sub} \subset S_{ind}$ as:
\begin{equation}
    S_{sub}=argmax_{i\in S_{ind}}(cos(q, e_s^i)), s.t. |S_{sub}|=k
\end{equation}
The final summary is obtained by concatenating the sentences associated with such indices, that is:
\begin{equation}
    \hat{y}=concat(s_i) \forall i \in S_{sub}
\end{equation}

\subsection{Evaluation}
\subsubsection{Aspect-Based Summarization Evaluation}
Following recent research in the field of summarization evaluation, we use the ChatGPT-RTS \cite{shen-etal-2023-large} for evaluation. This metric uses the powerful ChatGPT LLM (i.e. GPT 3.5) as an evaluator, by framing the evaluation task as a question concerning the property of the summaries with respect to 4 key attributes individuated by \citet{10.1162/tacl_a_00362}: coherence, consistency, fluency, and relevance. For each reference summary, paragraphs, and topic triplet, ChatGPT is given the definition of the dimension to evaluate as well as the triplet and asked to output a score from 1 to 5, together with an explanation for such a decision. We introduced a key modification to the relevance definition in the prompt to include the target topic so that, with minimal modification, the final score also takes into consideration the target aspect. In appendix \ref{sec:appendixA} we illustrate in more detail the prompts fed to ChatGPT for performing the evaluation, as well as the correlation with human judgment and comparison with other metrics.

\vspace{-0.5em}
\subsubsection{Retrieval Evaluation}
To assess how successful different encoders are in retrieving the correct paragraphs in the RAG setting, we use the Mean Reciprocal Rank (MRR) metric, an information retrieval metric that considers how high in a ranked list the retriever can place the correct item (in our case the correct paragraph) \cite{radev-etal-2002-evaluating}.  

In our case, we set the hyperparameter of MRR to 10, meaning that we consider the first 10 items as scored by the retriever as the limit beyond which we consider the retriever to have failed (leading to MRR@10 equals 0).
\vspace{-0.5em}
\subsection{Energy Consumption and Efficiency Re-Weighting}
The Carburacy method was proposed to account for efficiency in summarization evaluation, by re-weighting the ROUGE metric for summarization with the cost for running the model $C=E*D$, where $E$ is the cost of a single example measured as the kg of $CO_2$ emitted by summarization models and $D$ is the dataset size \cite{Moro_Ragazzi_Valgimigli_2023}. The re-weighting formula is then applied as:
\begin{equation}
    \gamma=\frac{e^{log_{\alpha}R}}{1+C*\beta}
    \label{eq:carburacy}
\end{equation}
with $R$ being the effectiveness score (i.e. the initial summarization metric) and $\alpha=10$ and $\beta=100$ following the original work. The authors further divided the costs in inference and training costs, but in our unsupervised setting just the first applies.

In applying the Carburacy re-weighting scheme to our context we took into account the fact that LLMs can lead to very different outcomes in terms of summaries length and this has an effect on the cost $C$ as longer sequences will lead to higher consumption in the auto-regressive setting of decoder-only modern LLMs. In our case, we want to isolate the cost of each LLM as a function solely of its architecture, rather than of its output. Therefore, we compute equation \ref{eq:carburacy} by setting $D=1$ and $E$ such that:
\begin{equation}
    E=Emission(LLM_{stop:k}(prompt_{fix}))
    \label{eq:llmcarbon}
\end{equation}
Where $prompt_{fix}$ is a fixed prompt for each system and $Emission$ is the function computing $CO_2$ emissions. The key of the above modification is represented by $LLM_{stop:k}$ which we define as a constrained generation from the given system, where the generation stops automatically at a token number $k$ which we set to 10. This way, each LLM receive a prompt of same input and output a same-length output, and by keeping these factors constant we assure to measure just differences in emissions caused by structural differences between LLMs (e.g. number of parameters).

When applying Carburacy to the extractive baselines and to the RAG models, instead, we simply apply equation \ref{eq:carburacy} with the cost of encoding $prompt_{fix}$ in the first case and with the cost of encoding the entire dataset $D$ in the latter. In the retrieval experiments, we empirically set $\beta=10000$ to account for the difference in emission scale.

We measure $CO_2$ levels with the codecarbon python library\footnote{https://codecarbon.io/}, leveraging CPU as well as GPU energy consumption.

\begin{table}
    \centering
    \small
    \begin{tabular}{l|c|c|c}
         Feature & AR5 & AR6 & All\\
         \hline
         Summaries & 70 & 70 & 140\\
         Paragraphs & 34 & 38 & 72 \\
         Summary Topics & 63 & 70 & 133 \\
         Summary Section Headers & 4 & 3 & 7 \\
         Summary Sub-Section Headers & 17 & 18 & 35\\
         Paragraphs Section Headers & 34 & 38 & 72 \\
    \end{tabular}
    \caption{Statistics of our IPCC-Sum dataset. For all features, we report the number of unique occurrences for the different subsets (AR5 and AR6), as well as for the whole dataset. It can be noticed how many topics are repeated in different summaries.}
\label{tab:1}
\end{table}
\vspace{-0.5em}
\section{Data}
\vspace{-0.5em}
For the purpose of this work, we have collected and released the SumIPCC dataset, comprising 140 topic-annotated summaries and relative paragraphs from climate change reports. We used two reports from the authoritative Intergovernmental Panel on Climate Change (IPCC) as a data source. The reports we used are the synthesis reports AR5 \cite{ipcc2014} and AR6 \cite{ipcc2023} for two separate years, 2014 and 2023, which collected the contributions of different working groups on a variety of topics related to climate change and linked policies.
The two reports were chosen among the IPCC synthesis report collections as they both include accompanying publications named Summary for Policy-Makers \cite{ipccsum2014, ipccsum2023}, which include short summaries related to specific topics and referring to paragraphs in the respective synthesis reports. Each summary includes the main highlights with regard to a specific topic as discussed in the report and it might refer to multiple paragraphs in the original report, in case the specific topic is treated in different parts of the report. 



On occasions, we observed summaries that were too broad in scope, referring to many different long paragraphs, but comprising just a few lines on a broad topic: we filtered out these cases. The final result is a dataset comprising 140 paragraph-summary pairs with associated topics, which we manually annotated to be as precise as possible. Paragraphs and section headers from the Summary for Policy-Makers could also have been used to annotate the summaries, but they were ambiguous as they grouped different summaries; they are also included as features in the dataset, even though we don't explore their use in this work. As we will see, however, there are a number of summaries sharing the same topic but in different contexts and future work might include additional information to better disambiguate these cases, especially in the RAG context. Table \ref{tab:1} shows the features from the collected dataset and their occurrences, while
Appendix \ref{sec:appendixE} includes additional information.

\begin{table}
    \centering
    \small
    \begin{tabular}{l|c|c|c}
         Model & Billions of Parameters & $C$ &\\
         \hline
         Qwen 0.5B & 0.5 & 4.06e-05 \\
         Qwen 1.8B & 1.8 & 4.19e-05 \\
         Qwen 4B & 4 & 5.28e-05 \\
         Qwen 7B & 7 & 5.63e-05 \\
         Gemma 2B & 2 & 4.41e-05 \\
         Gemma 7B & 7 & 6.41e-05 \\
         Phi 3 & 3.8 & 5.30e-05 \\
         Llama 3 & 8 & 6.20e-05 \\
         Mistral & 7 & 6.03e-05 \\
         \hline
         ChatGPT & \(\sim \) 175 & \(\sim \) 3.86e-03 \\
         GPT4 & \(\sim \) 175 & \(\sim \) 3.86e-03\\
         \hline
         MPNet & 0.11& 1.65e-07
    \end{tabular}
    \caption{Number of parameters and estimated energy cost $C$ for the ABS models. In every case, we used the conventional abbreviated notation, e.g., e-05 to signify a multiplier of $10^{-5}$ for the given value. Model size does not perfectly correlate with energy consumption, as different architectures might have different efficiency.}
\label{tab:2}
\end{table}
\vspace{-0.5em}
\section{Experimental Setup}
\subsection{LLMs and Extractive Baselines}
\label{sec:exp}
We compare recent and popular LLMs: 9 open-source SLMs and 2 big, proprietary LLMs. For the SLMs, there is no single definition of how small a model should be to be considered such, therefore we impose a hardware constraint to choose the models, namely to be able to fit in a single NVIDIA® Tesla T4 GPU with 16GB of memory: to achieve this, we have then selected models up to 8 billion parameters, while using 4-bit quantization on all the models from this category; the effect of the quantization has been shown to be negligible in most cases \cite{Yao_Wu_Li_Youn_He_2024}. The SLMs we used are: Qwen 1.5 (Qwen) 0.5B, 1.8B, 4B and 7B \cite{bai2023qwen}, Gemma 1.1 (Gemma) 2B and 7B \cite{gemmateam2024gemma}, Phi 3 \cite{abdin2024phi3}, Llama 3 8B (Llama 3) \cite{llama3} and Mistral v0.2 7B (Mistral) \cite{jiang2023mistral}. In every case, we have used the instruction-tuned versions of the models: we give additional details about the models' source and run time in Appendix \ref{sec:appendixF}. 

To compare the performance of SLMs with bigger LLMs, we compare them with the state-of-the-art GPT4 \cite{openai2024gpt4} and its earlier version, ChatGPT \cite{brown2020language}; no public information about the quantization settings nor the model size exist for the two models, but table \ref{tab:2} includes estimates on size and energy cost $C$ for these models together with the actual models size and cost for the small-sized LLMs. We computed $C$ as per equation \ref{eq:llmcarbon}, while we report a rough estimate of the sizes of GPT4 and ChatGPT by equating them to the size of the related model GPT3 \cite{brown2020language} and we estimated their cost $C$ by multiplying the cost of Gemma 2B for the module of the respective model parameters; this is indeed a very rough estimate, but it should give a good approximation of the scale of difference between small-sized LLMs and bigger, state-of-the-art ones. Finally, for the extractive baselines we have used the all-mpnet-base-v2 (MPNet) model, further described in the next section. Also for this models, we include the energy cost $C$ in table \ref{tab:2}.

\begin{table}
    \centering
    \small
    \begin{tabular}{l|c|c|c}
         Model & Billions of Parameters & $C$ &\\
         \hline
         DistilRoB & 0.08 & 4.06e-05 \\
         MPNet & 0.11 & 4.19e-05 \\
         MiniLM & 0.2 & 4.42e-10 \\
         GTR & 1.2 & 5.63e-05 \\
         ST5 & 1.2 & 4.41e-05 \\
         GTE & 0.44 & 6.41e-05 \\
    \end{tabular}
    \caption{Number of parameters and estimated energy cost $C$ for the text encoders used as zero-shot retrievers in our RAG experiments.}
\label{tab:3}
\end{table}
\vspace{-0.5em}
\subsection{Retrieval and Extractive Models}
To choose the zero-shot text retrieval models for the RAG experiments, we have mostly drawn from the top open-source systems from the MTEB benchmarks evaluating out-of-the-box text embedding systems \cite{muennighoff-etal-2023-mteb}. At the same time, we have included the same hardware constraints explained in section \ref{sec:exp} to limit our choice to relatively small-sized encoders. The final models we used in the RAG setting, then, are: all-mpnet-base-v2 (MPNet), an encoder based on the MPNet architecture \cite{10.5555/3495724.3497138} and on the sentence transformers framework \cite{reimers-gurevych-2019-sentence} to be highly performative in a variety of sentence encoding tasks, all-distilroberta-v1 (DistilRoB), a distilled version of RoBERTa \cite{liu2019roberta} trained similarly to MPNet, all-MiniLM-L12-v2 (MiniLM), a small and extremely efficient transformer encoder \cite{gu2024minillm} further fine-tuned similarly to MPNet, gtr-t5-xl (GTR) \cite{ni-etal-2022-large} and sentence-t5-xl (ST5) \cite{ni-etal-2022-sentence}, two sentence encoders both based on the encoder part of the T5 architecture \cite{T5} but fine-tuned on different datasets for text retrieval, and gte-large-en-v1.5 (GTE) \cite{li2023general}, a transformer encoder trained with multi-stage contrastive learning.

Table \ref{tab:3} shows the number of parameters for this set of models, together with the energy cost $C$ computed as described in the methodology section.

\begin{figure}
    \centering
    \includegraphics[height=11em, width=16em]{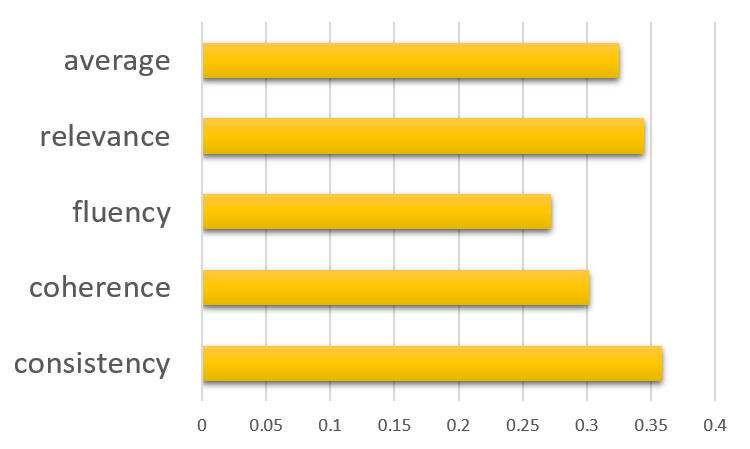}
    \caption{Pearsons' correlation between the metrics' aspects and energy consumption.}
    \label{fig:emission_correlation}
\end{figure}
\vspace{-0.5em}
\section{Experiments}
\begin{table*}[th]
    \centering
    \small
    \begin{tabular}{l|c|c|c|c|c}
         Model & Consistency $\uparrow$ & Coherence $\uparrow$ & Fluency $\uparrow$ & Relevance $\uparrow$ & Average $\uparrow$\\
         \hline
         \hline
         Qwen 0.5B & 4.52*&4.33*&4.41*&4.06*&4.33* \\
         Qwen 1.8B & 4.89&4.83&4.88&4.79&4.85 \\
         Qwen 4B & 4.75*&4.84&4.91&4.56*&4.77 \\
         Qwen 7B & 4.84&4.94&4.9&4.74&4.86 \\
         Gemma 2B & 4.86&4.86&4.96&4.71&4.85 \\
         Gemma 7B & 4.85&4.94&4.99&\textbf{4.81}&4.9 \\
         Phi 3 & 4.84&4.92&4.94&4.74&4.86 \\
         Llama 3 & 4.82&4.84&4.91&4.74&4.83 \\
         Mistral & 4.78*&4.84&4.95&4.6&4.79 \\
         \hline
         ChatGPT & \textbf{4.94}&\textbf{4.96}&\textbf{4.98}&4.79&\textbf{4.91}\\
         GPT4 & 4.83&4.89&4.96&\textbf{4.81}&4.89\\
         \hline
         MPNet & 4.44*&3.03*&3.45*&4.15*&3.77*
    \end{tabular}
    \caption{Summarization results for all dimensions obtained by evaluating our models with the ChatGPT-RTS metric. Asterisks indicate that the results are significantly worse than the best model (i.e. ChatGPT).}
\label{tab:4}
\end{table*}

\subsection{SLMs Evaluation}
Table \ref{tab:4} shows the results obtained by running and comparing to reference summaries our SLMs and baselines over the SumIPCC dataset with the ground truth paragraphs for each reference summary (i.e. without RAG). The results clearly highlight a very good performance on behalf of most SLMs and LLMs, whereas the extractive baselines show inferior performance for all the given evaluation dimensions; such a difference is statistically significant ($p<0.05$)\footnote{A two sample T-test was performed to determine significance, with the best results (i.e. ChatGPT) serving as the control group for comparison.} and it confirms the superiority of LLMs of any size to the simple extractive models (Q1). It is interesting to notice, however, that the performance of the extractive method is generally good in absolute terms for the relevance and consistency dimensions, highlighting the style of this dataset, where many exact lines from the target paragraphs are present in the reference summaries (see appendix \ref{sec:appendixA} and appendix \ref{sec:appendixE}).

When comparing SLMs with the LLMs baselines, we can observe some striking results in that the ChatGPT baseline appears to be the best-performing system overall, even more so than the superior GPT4 baseline. This apparently counter-intuitive result can, however, be explained by three factors: first, as the metric we use is based on ChatGPT itself it might show a bias in favor of the model, as observed in previous studies \cite{panickssery2024llm}, second, the reliability of the metric in the context of high-quality summaries is generally lower \cite{shen-etal-2023-large}, and third, ChatGPT is not significantly better than GPT4 in any evaluation dimension. These points also apply to most SLMs. More recent and relatively more powerful SLMs like Llama 3, in fact, appear to be worse than other models like ChatGPT itself, but ultimately the difference is statistically insignificant, rather indicating that most SLMs and LLMs perform similarly in our context.
SLMs, then, can be as effective as larger LLMs for our task (Q2).

\begin{figure}
    \centering
    \includegraphics[height=17em, width=17em]{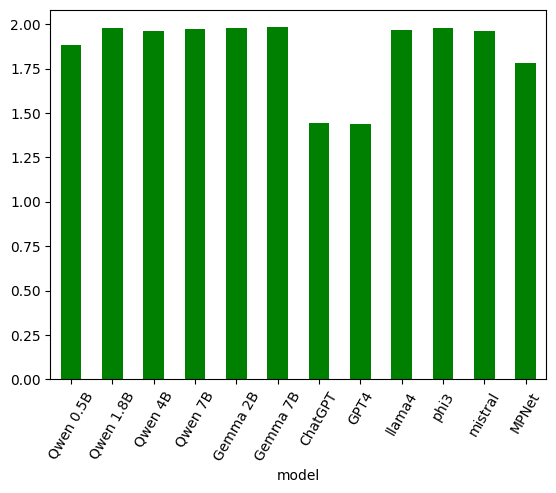}
    \caption{ChatGPT RTS Average scores re-weighted via Carburacy.}
    \label{fig:reranking1}
\end{figure}
Moreover, figure \ref{fig:emission_correlation} shows how the energy consumption shows a weak, but relevant correlation with LLMs performance on each dimension. A key driver of this correlation is the poor performance of Qwen 0.5B, suggesting that there is a threshold under which model size severely impacts the capacity of SLMs to perform this task. The updated ranking of models in figure \ref{fig:reranking1} using the Carburacy technique, however, shows how on certain occasions, notably that of Qwen 1.8B, very small SLMs can perform similarly to larger ones. The re-ranking confirms once more that most SLMs perform similarly, and that are generally better than very small LLMs (Qwen 0.5B) and then the extractive baseline. It follows, that ChatGPT and GPT4 are actually the worst models when considering the efficiency/effectiveness trade-off because the increase in energy consumption is not justified by a relevant increase in the models' performance.

\begin{table*}[th]
    \centering
    \small
    \begin{tabular}{l|c|c|c|c|c}
         Model & Consistency $\uparrow$ & Coherence $\uparrow$ & Fluency $\uparrow$ & Relevance $\uparrow$ & Average $\uparrow$\\
         \hline
         \hline
         Qwen 1.8B & 3.66&\textbf{4.36}&4.24&3.11&3.84 \\
         Gemma 2B & 3.21*&3.81*&3.67*&3.21&3.48* \\
         Phi 3 & 3.32*&3.82*&3.74*&3.23&3.53* \\
         Llama 3 & \textbf{3.76}&4.27&\textbf{4.44}&\textbf{3.26}&\textbf{3.93} \\
         Mistral & 3.02*&3.61*&3.56*&3.02&3.30* \\
         \hline
         ChatGPT & 3.24*&3.81*&3.52*&2.96&3.38*\\
         \hline
         MPNet & 2.68*&2.39*&2.5*&2.36*&2.48*
    \end{tabular}
    \caption{Summarization results for all dimensions obtained by evaluating our models with the ChatGPT-RTS metric on the retrieved passages. Asterisks indicate results that are significantly worse than the best model (i.e. Llama 3).}
\label{tab:4}
\end{table*}
\vspace{-0.5em}

\begin{figure}[!thb]
    \centering
    \includegraphics[width=17em,height=15em]{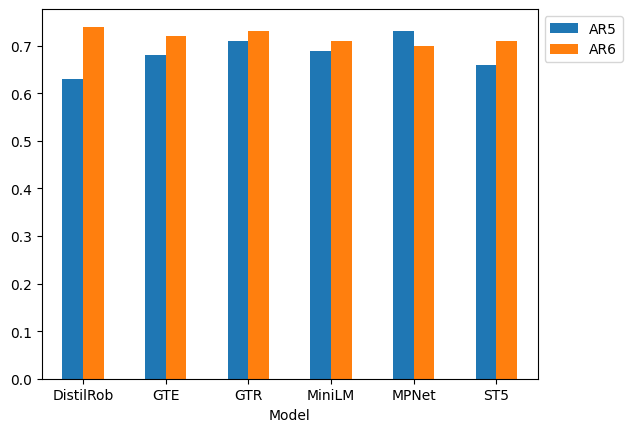}
    \caption{Retrieval results in terms of MRR@10 metric re-weighted via the Carburacy method.}
    \label{fig:retrieval}
\end{figure}

\subsection{RAG Evaluation}

Figure \ref{fig:retrieval} shows the results of using different retrieval models on the two subsets of our dataset, separately. It can be seen how also in this case most models perform similarly and, applying the Carburacy method to re-weight the MRR@10 score, this leads to comparatively smaller models being the best choice to perform the retrieval in our context.

\begin{figure}
    \centering
    \includegraphics[width=15em,height=15em]{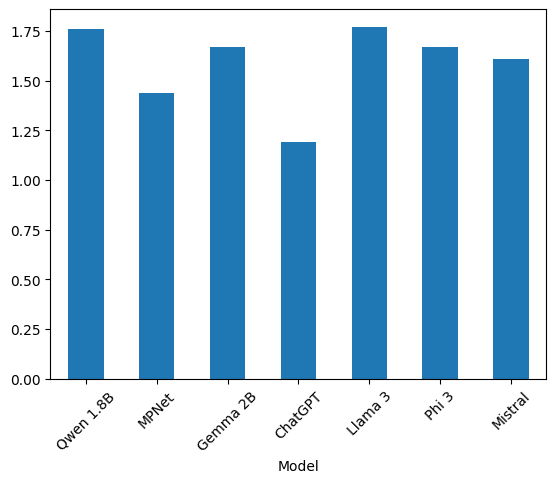}
    \caption{ChatGPT RTS Average scores for the RAG experiment re-weighted via the Carburacy method.}
    \label{fig:rag_carb}
\end{figure}

Having identified the best retrieval models for both subsets of our dataset, we employ them to retrieve the top 2 paragraphs for each query and then we employ the method described in section 3.2 to generate the summaries. In this case, we have used just the best models for each family, as indicated by results in table \ref{tab:4}. It is interesting to notice how this time the results from different models are more spread, highlighting more significant differences individuated by our metric in this more challenging scenario. This is in line with what was previously observed for the same metric, as using ChatGPT to evaluate ABS has been shown to be more accurate and more confident about its own decision when the difference in the quality of the generated summaries is substantial \cite{shen-etal-2023-large}. The fact of using two paragraphs that might not be the correct ones as input to be summarized according to a specific topic, in fact, seems to have an effect on all dimensions, not only on the relevance one (which presents the biggest overall drop in performance, as it could have been expected). This evidence suggests that our task in a RAG setting is indeed a more challenging task, which requires further investigation both in terms of the retrieval model being used and in terms of the summarization model. Different LLMs, in fact, appear to be more capable of dealing with heterogeneous information and filter out irrelevant information, while maintaining good coherence, fluency, and consistency with the input paragraphs (more qualitative examples under this respect are presented in \ref{sec:appendixD}). Because of this, in this context the choice of the model appears to be relevant, with Llama 3 performing significantly better than most other models, in line with the models' performance on existing benchmarks \cite{llama3}. Interestingly it can be seen how the much smaller Qwen 1.8B, however, performs similarly to Llama 3 and this leads to the model being ranked as good as the latter in the re-weighted results using Carburacy, shown in figure \ref{fig:rag_carb}. This last evidence shows once more that smaller LLMs can perform as well as bigger ones in our context and this might be because of a variety of reasons including training data, stochasticity, and prompt preferences: the inclusion of efficiency in the evaluation framework allowed to identify models with a smaller energy-cost, while leading to a drop in performance which is minimal or even not significant.

\section{Conclusion and Future Directions}
In this work, we have investigated the use of LLMs and SLMs for ABS in the context of climate change reports, showing how the task can be virtually solved by such models when considering ground truth paragraphs. Apart from the task itself, which has a variety of uses in policy-making and education, our aim was also that of evaluating whether smaller, more efficient LLMs (i.e. SLMs) can lead to comparable results to bigger one in a task in which LLMs are extremely capable. The results indeed confirmed that SLMs are a valid alternative to bigger LLMs, especially in the easier scenario in which ground truth paragraphs were provided. 

When we turned to the RAG scenario, instead, it could be seen that the task at hand became more challenging, while the difference in the models we used appeared to be more significant. Also in this case, however, the smallest model performed comparably with the best-performing one and, even though this might be due to various things not reflecting a more general equivalence, the evidence suggests, at least, that smaller models can be a valid alternative also in more challenging cases.

Finally, we release our dataset and this can lead to many interesting research directions. Specifically, future research could explore the RAG setting further by incorporating more fine-grained information during retrieval (e.g. section and/or paragraph titles, which are included in the dataset) and fine-tune SLMs on the small available data to test the ability of such models to learn from small data. We leave these directions open for future research.


\section{Limitations}
Our work deals with the use of SLMs for ABS and has shown that they often perform similarly to larger LLMs in our context. Given the specific domain of application (i.e. climate change reports), however, we are limited to a small size dataset, which in turn increases results' variability.
Another limitation of our work involve the evaluation metric, which includes a number of problems such as having around 80\% agreement with human judgement, as shown in appendix \ref{sec:appendixA}: this value is relatively high for summarization metrics, but it is still low enough to represent a significant limitation in terms of how much we can trust the metric itself in certain cases. Other evaluation limitations include the fact that our metric has been shown to correlate less with human judgement when dealing with high-performing systems (which is our case in the first experiment using ground truth paragraphs) and the already noticed fact that the metric appears to be biased towards certain LLMs (i.e. ChatGPT).

Finally, there is initial evidence that the aspects we have evaluated for each sample in our dataset might be too broad. Future research might consider using the additional features we provided in the released dataset in order to better define the aspect on which the summarization models should focus.

\section{Ethical Considerations}
Using LLMs and SLMs to summarize climate change reports raises several ethical considerations:

1) \textbf{Accuracy and Reliability}.
If inaccurate or misleading summaries are produced by LLMs, this could potentially misinform stakeholders and the public, leading to poor decision-making. Therefore, it is essential to have a human-in-the-loop approach in double-checking the produced summaries.

2) \textbf{Transparency and Accountability}.
LLMs are black-box and therefore are not transparent nor accountable in terms of what output they produce. Notwithstanding the de-biasing and alignment with human preferences that the systems we used undertook, the reasons why such models produced certain summaries remain opaque.

3) \textbf{Accessibility and Inclusivity}.
The use of LLMs require access to resources that might not be widely available in less developed countries and poorly funded institutions and, therefore, these could lead to problem of inclusivity and reduced access to our tool for policy-makers and educators from such background.

\bibliography{custom}

\begin{thebibliography}{49}
\providecommand{\natexlab}[1]{#1}

\bibitem[{Abdin et~al.(2024)Abdin, Jacobs, Awan, Aneja, Awadallah, Awadalla, Bach, Bahree, Bakhtiari, Behl, Benhaim, Bilenko, Bjorck, Bubeck, Cai, Mendes, Chen, Chaudhary, Chopra, Giorno, de~Rosa, Dixon, Eldan, Iter, Garg, Goswami, Gunasekar, Haider, Hao, Hewett, Huynh, Javaheripi, Jin, Kauffmann, Karampatziakis, Kim, Khademi, Kurilenko, Lee, Lee, Li, Liang, Liu, Lin, Lin, Madan, Mitra, Modi, Nguyen, Norick, Patra, Perez-Becker, Portet, Pryzant, Qin, Radmilac, Rosset, Roy, Ruwase, Saarikivi, Saied, Salim, Santacroce, Shah, Shang, Sharma, Song, Tanaka, Wang, Ward, Wang, Witte, Wyatt, Xu, Xu, Yadav, Yang, Yang, Yu, Zhang, Zhang, Zhang, Zhang, Zhang, Zhang, Zhang, and Zhou}]{abdin2024phi3}
Marah Abdin, Sam~Ade Jacobs, Ammar~Ahmad Awan, Jyoti Aneja, Ahmed Awadallah, Hany Awadalla, Nguyen Bach, Amit Bahree, Arash Bakhtiari, Harkirat Behl, Alon Benhaim, Misha Bilenko, Johan Bjorck, Sébastien Bubeck, Martin Cai, Caio César~Teodoro Mendes, Weizhu Chen, Vishrav Chaudhary, Parul Chopra, Allie~Del Giorno, Gustavo de~Rosa, Matthew Dixon, Ronen Eldan, Dan Iter, Amit Garg, Abhishek Goswami, Suriya Gunasekar, Emman Haider, Junheng Hao, Russell~J. Hewett, Jamie Huynh, Mojan Javaheripi, Xin Jin, Piero Kauffmann, Nikos Karampatziakis, Dongwoo Kim, Mahoud Khademi, Lev Kurilenko, James~R. Lee, Yin~Tat Lee, Yuanzhi Li, Chen Liang, Weishung Liu, Eric Lin, Zeqi Lin, Piyush Madan, Arindam Mitra, Hardik Modi, Anh Nguyen, Brandon Norick, Barun Patra, Daniel Perez-Becker, Thomas Portet, Reid Pryzant, Heyang Qin, Marko Radmilac, Corby Rosset, Sambudha Roy, Olatunji Ruwase, Olli Saarikivi, Amin Saied, Adil Salim, Michael Santacroce, Shital Shah, Ning Shang, Hiteshi Sharma, Xia Song, Masahiro Tanaka, Xin Wang, Rachel
  Ward, Guanhua Wang, Philipp Witte, Michael Wyatt, Can Xu, Jiahang Xu, Sonali Yadav, Fan Yang, Ziyi Yang, Donghan Yu, Chengruidong Zhang, Cyril Zhang, Jianwen Zhang, Li~Lyna Zhang, Yi~Zhang, Yue Zhang, Yunan Zhang, and Xiren Zhou. 2024.
\newblock \href {https://arxiv.org/abs/2404.14219} {Phi-3 technical report: A highly capable language model locally on your phone}.
\newblock \emph{Preprint}, arXiv:2404.14219.

\bibitem[{Ahuja et~al.(2022)Ahuja, Xu, Gupta, Horecka, and Durrett}]{ahuja-etal-2022-aspectnews}
Ojas Ahuja, Jiacheng Xu, Akshay Gupta, Kevin Horecka, and Greg Durrett. 2022.
\newblock \href {https://doi.org/10.18653/v1/2022.acl-long.449} {{ASPECTNEWS}: Aspect-oriented summarization of news documents}.
\newblock In \emph{Proceedings of the 60th Annual Meeting of the Association for Computational Linguistics (Volume 1: Long Papers)}, pages 6494--6506, Dublin, Ireland. Association for Computational Linguistics.

\bibitem[{Bai et~al.(2023)Bai, Bai, Chu, Cui, Dang, Deng, Fan, Ge, Han, Huang, Hui, Ji, Li, Lin, Lin, Liu, Liu, Lu, Lu, Ma, Men, Ren, Ren, Tan, Tan, Tu, Wang, Wang, Wang, Wu, Xu, Xu, Yang, Yang, Yang, Yang, Yao, Yu, Yuan, Yuan, Zhang, Zhang, Zhang, Zhang, Zhou, Zhou, Zhou, and Zhu}]{bai2023qwen}
Jinze Bai, Shuai Bai, Yunfei Chu, Zeyu Cui, Kai Dang, Xiaodong Deng, Yang Fan, Wenbin Ge, Yu~Han, Fei Huang, Binyuan Hui, Luo Ji, Mei Li, Junyang Lin, Runji Lin, Dayiheng Liu, Gao Liu, Chengqiang Lu, Keming Lu, Jianxin Ma, Rui Men, Xingzhang Ren, Xuancheng Ren, Chuanqi Tan, Sinan Tan, Jianhong Tu, Peng Wang, Shijie Wang, Wei Wang, Shengguang Wu, Benfeng Xu, Jin Xu, An~Yang, Hao Yang, Jian Yang, Shusheng Yang, Yang Yao, Bowen Yu, Hongyi Yuan, Zheng Yuan, Jianwei Zhang, Xingxuan Zhang, Yichang Zhang, Zhenru Zhang, Chang Zhou, Jingren Zhou, Xiaohuan Zhou, and Tianhang Zhu. 2023.
\newblock \href {https://arxiv.org/abs/2309.16609} {Qwen technical report}.
\newblock \emph{Preprint}, arXiv:2309.16609.

\bibitem[{Bhaskar et~al.(2023)Bhaskar, Fabbri, and Durrett}]{bhaskar-etal-2023-prompted}
Adithya Bhaskar, Alex Fabbri, and Greg Durrett. 2023.
\newblock \href {https://doi.org/10.18653/v1/2023.findings-acl.591} {Prompted opinion summarization with {GPT}-3.5}.
\newblock In \emph{Findings of the Association for Computational Linguistics: ACL 2023}, pages 9282--9300, Toronto, Canada. Association for Computational Linguistics.

\bibitem[{Biester et~al.(2022)Biester, Demszky, Jin, Sachan, Tetreault, Wilson, Xiao, and Zhao}]{nlp4pi-2022-nlp}
Laura Biester, Dorottya Demszky, Zhijing Jin, Mrinmaya Sachan, Joel Tetreault, Steven Wilson, Lu~Xiao, and Jieyu Zhao, editors. 2022.
\newblock \href {https://aclanthology.org/2022.nlp4pi-1.0} {\emph{Proceedings of the Second Workshop on NLP for Positive Impact (NLP4PI)}}. Association for Computational Linguistics, Abu Dhabi, United Arab Emirates (Hybrid).

\bibitem[{Brown et~al.(2020)Brown, Mann, Ryder, Subbiah, Kaplan, Dhariwal, Neelakantan, Shyam, Sastry, Askell, Agarwal, Herbert-Voss, Krueger, Henighan, Child, Ramesh, Ziegler, Wu, Winter, Hesse, Chen, Sigler, Litwin, Gray, Chess, Clark, Berner, McCandlish, Radford, Sutskever, and Amodei}]{brown2020language}
Tom~B. Brown, Benjamin Mann, Nick Ryder, Melanie Subbiah, Jared Kaplan, Prafulla Dhariwal, Arvind Neelakantan, Pranav Shyam, Girish Sastry, Amanda Askell, Sandhini Agarwal, Ariel Herbert-Voss, Gretchen Krueger, Tom Henighan, Rewon Child, Aditya Ramesh, Daniel~M. Ziegler, Jeffrey Wu, Clemens Winter, Christopher Hesse, Mark Chen, Eric Sigler, Mateusz Litwin, Scott Gray, Benjamin Chess, Jack Clark, Christopher Berner, Sam McCandlish, Alec Radford, Ilya Sutskever, and Dario Amodei. 2020.
\newblock \href {https://arxiv.org/abs/2005.14165} {Language models are few-shot learners}.
\newblock \emph{Preprint}, arXiv:2005.14165.

\bibitem[{Coavoux et~al.(2019)Coavoux, Elsahar, and Gall{\'e}}]{coavoux-etal-2019-unsupervised}
Maximin Coavoux, Hady Elsahar, and Matthias Gall{\'e}. 2019.
\newblock \href {https://doi.org/10.18653/v1/D19-5405} {Unsupervised aspect-based multi-document abstractive summarization}.
\newblock In \emph{Proceedings of the 2nd Workshop on New Frontiers in Summarization}, pages 42--47, Hong Kong, China. Association for Computational Linguistics.

\bibitem[{Faiz et~al.(2024)Faiz, Kaneda, Wang, Osi, Sharma, , and Jiang}]{2024ICLR:LLMCarbon}
Ahmad Faiz, Sotaro Kaneda, Ruhan Wang, Rita Osi, Prateek Sharma, , and Lei Jiang. 2024.
\newblock {LLMCarbon:} modeling the end-to-end carbon footprint of large language models.
\newblock In \emph{International Conference on Learning Representations (ICLR)}.

\bibitem[{Fraile-Hernandez and Pe{\~n}as(2024)}]{fraile-hernandez-penas-2024-hamison}
Jesus~M. Fraile-Hernandez and Anselmo Pe{\~n}as. 2024.
\newblock \href {https://aclanthology.org/2024.case-1.10} {{HAM}i{S}o{N}-generative at {C}limate{A}ctivism 2024: Stance detection using generative large language models}.
\newblock In \emph{Proceedings of the 7th Workshop on Challenges and Applications of Automated Extraction of Socio-political Events from Text (CASE 2024)}, pages 79--84, St. Julians, Malta. Association for Computational Linguistics.

\bibitem[{Frermann and Klementiev(2019)}]{frermann-klementiev-2019-inducing}
Lea Frermann and Alexandre Klementiev. 2019.
\newblock \href {https://doi.org/10.18653/v1/P19-1630} {Inducing document structure for aspect-based summarization}.
\newblock In \emph{Proceedings of the 57th Annual Meeting of the Association for Computational Linguistics}, pages 6263--6273, Florence, Italy. Association for Computational Linguistics.

\bibitem[{Gu et~al.(2024)Gu, Dong, Wei, and Huang}]{gu2024minillm}
Yuxian Gu, Li~Dong, Furu Wei, and Minlie Huang. 2024.
\newblock \href {https://arxiv.org/abs/2306.08543} {Minillm: Knowledge distillation of large language models}.
\newblock \emph{Preprint}, arXiv:2306.08543.

\bibitem[{Hayashi et~al.(2021)Hayashi, Budania, Wang, Ackerson, Neervannan, and Neubig}]{10.1162/tacl_a_00362}
Hiroaki Hayashi, Prashant Budania, Peng Wang, Chris Ackerson, Raj Neervannan, and Graham Neubig. 2021.
\newblock \href {https://doi.org/10.1162/tacl_a_00362} {{WikiAsp: A Dataset for Multi-domain Aspect-based Summarization}}.
\newblock \emph{Transactions of the Association for Computational Linguistics}, 9:211--225.

\bibitem[{IPCC(2014{\natexlab{a}})}]{ipccsum2014}
IPCC. 2014{\natexlab{a}}.
\newblock Climate change 2014: Summary for policy-makers.
\newblock Technical report, IPCC.

\bibitem[{IPCC(2014{\natexlab{b}})}]{ipcc2014}
IPCC. 2014{\natexlab{b}}.
\newblock Climate change 2014: Synthesis report.
\newblock Technical report, IPCC.

\bibitem[{IPCC(2023{\natexlab{a}})}]{ipccsum2023}
IPCC. 2023{\natexlab{a}}.
\newblock Climate change 2023: Summary for policy-makers.
\newblock Technical report, IPCC.

\bibitem[{IPCC(2023{\natexlab{b}})}]{ipcc2023}
IPCC. 2023{\natexlab{b}}.
\newblock Climate change 2023: Synthesis report.
\newblock Technical report, IPCC.

\bibitem[{Jeong and Lee(2024)}]{su16041640}
Nayoung Jeong and Jihwan Lee. 2024.
\newblock \href {https://doi.org/10.3390/su16041640} {An aspect-based review analysis using chatgpt for the exploration of hotel service failures}.
\newblock \emph{Sustainability}, 16(4).

\bibitem[{Jiang et~al.(2023)Jiang, Sablayrolles, Mensch, Bamford, Chaplot, de~las Casas, Bressand, Lengyel, Lample, Saulnier, Lavaud, Lachaux, Stock, Scao, Lavril, Wang, Lacroix, and Sayed}]{jiang2023mistral}
Albert~Q. Jiang, Alexandre Sablayrolles, Arthur Mensch, Chris Bamford, Devendra~Singh Chaplot, Diego de~las Casas, Florian Bressand, Gianna Lengyel, Guillaume Lample, Lucile Saulnier, Lélio~Renard Lavaud, Marie-Anne Lachaux, Pierre Stock, Teven~Le Scao, Thibaut Lavril, Thomas Wang, Timothée Lacroix, and William~El Sayed. 2023.
\newblock \href {https://arxiv.org/abs/2310.06825} {Mistral 7b}.
\newblock \emph{Preprint}, arXiv:2310.06825.

\bibitem[{Kiss and Strunk(2006)}]{punkt}
Tibor Kiss and Jan Strunk. 2006.
\newblock \href {https://doi.org/10.1162/coli.2006.32.4.485} {Unsupervised multilingual sentence boundary detection}.
\newblock \emph{Computational Linguistics}, 32:485--525.

\bibitem[{Li et~al.(2023)Li, Zhang, Zhang, Long, Xie, and Zhang}]{li2023general}
Zehan Li, Xin Zhang, Yanzhao Zhang, Dingkun Long, Pengjun Xie, and Meishan Zhang. 2023.
\newblock \href {https://arxiv.org/abs/2308.03281} {Towards general text embeddings with multi-stage contrastive learning}.
\newblock \emph{Preprint}, arXiv:2308.03281.

\bibitem[{Liu et~al.(2019)Liu, Ott, Goyal, Du, Joshi, Chen, Levy, Lewis, Zettlemoyer, and Stoyanov}]{liu2019roberta}
Yinhan Liu, Myle Ott, Naman Goyal, Jingfei Du, Mandar Joshi, Danqi Chen, Omer Levy, Mike Lewis, Luke Zettlemoyer, and Veselin Stoyanov. 2019.
\newblock \href {https://arxiv.org/abs/1907.11692} {Roberta: A robustly optimized bert pretraining approach}.
\newblock \emph{Preprint}, arXiv:1907.11692.

\bibitem[{Ma et~al.(2022)Ma, Pan, Rong, Qian, Tian, and Al-Nabhan}]{supervised-aspect-based}
Tinghuai Ma, Qian Pan, Huan Rong, Yurong Qian, Yuan Tian, and Najla Al-Nabhan. 2022.
\newblock \href {https://doi.org/10.1109/TCSS.2021.3088506} {T-bertsum: Topic-aware text summarization based on bert}.
\newblock \emph{IEEE Transactions on Computational Social Systems}, 9(3):879--890.

\bibitem[{Mazid and Zarnaz(2022)}]{fakenews-det}
Md~Abdullah~Al Mazid and Zaima Zarnaz. 2022.
\newblock \href {https://doi.org/10.1145/3542954.3542995} {Climate change myths detection using dynamically weighted ensemble based stance classifier}.
\newblock In \emph{Proceedings of the 2nd International Conference on Computing Advancements}, ICCA '22, page 277–283, New York, NY, USA. Association for Computing Machinery.

\bibitem[{Meddeb et~al.(2022)Meddeb, Ruseti, Dascalu, Terian, and Travadel}]{su141811724}
Paul Meddeb, Stefan Ruseti, Mihai Dascalu, Simina-Maria Terian, and Sebastien Travadel. 2022.
\newblock \href {https://doi.org/10.3390/su141811724} {Counteracting french fake news on climate change using language models}.
\newblock \emph{Sustainability}, 14(18).

\bibitem[{Meta(2024)}]{llama3}
Meta. 2024.
\newblock Introducing meta llama 3: The most capable openly available llm to date.

\bibitem[{Moro et~al.(2023)Moro, Ragazzi, and Valgimigli}]{Moro_Ragazzi_Valgimigli_2023}
Gianluca Moro, Luca Ragazzi, and Lorenzo Valgimigli. 2023.
\newblock \href {https://doi.org/10.1609/aaai.v37i12.26686} {Carburacy: Summarization models tuning and comparison in eco-sustainable regimes with a novel carbon-aware accuracy}.
\newblock \emph{Proceedings of the AAAI Conference on Artificial Intelligence}, 37(12):14417--14425.

\bibitem[{Muennighoff et~al.(2023)Muennighoff, Tazi, Magne, and Reimers}]{muennighoff-etal-2023-mteb}
Niklas Muennighoff, Nouamane Tazi, Loic Magne, and Nils Reimers. 2023.
\newblock \href {https://doi.org/10.18653/v1/2023.eacl-main.148} {{MTEB}: Massive text embedding benchmark}.
\newblock In \emph{Proceedings of the 17th Conference of the European Chapter of the Association for Computational Linguistics}, pages 2014--2037, Dubrovnik, Croatia. Association for Computational Linguistics.

\bibitem[{Ni et~al.(2022{\natexlab{a}})Ni, Hernandez~Abrego, Constant, Ma, Hall, Cer, and Yang}]{ni-etal-2022-sentence}
Jianmo Ni, Gustavo Hernandez~Abrego, Noah Constant, Ji~Ma, Keith Hall, Daniel Cer, and Yinfei Yang. 2022{\natexlab{a}}.
\newblock \href {https://doi.org/10.18653/v1/2022.findings-acl.146} {Sentence-t5: Scalable sentence encoders from pre-trained text-to-text models}.
\newblock In \emph{Findings of the Association for Computational Linguistics: ACL 2022}, pages 1864--1874, Dublin, Ireland. Association for Computational Linguistics.

\bibitem[{Ni et~al.(2022{\natexlab{b}})Ni, Qu, Lu, Dai, Hernandez~Abrego, Ma, Zhao, Luan, Hall, Chang, and Yang}]{ni-etal-2022-large}
Jianmo Ni, Chen Qu, Jing Lu, Zhuyun Dai, Gustavo Hernandez~Abrego, Ji~Ma, Vincent Zhao, Yi~Luan, Keith Hall, Ming-Wei Chang, and Yinfei Yang. 2022{\natexlab{b}}.
\newblock \href {https://doi.org/10.18653/v1/2022.emnlp-main.669} {Large dual encoders are generalizable retrievers}.
\newblock In \emph{Proceedings of the 2022 Conference on Empirical Methods in Natural Language Processing}, pages 9844--9855, Abu Dhabi, United Arab Emirates. Association for Computational Linguistics.

\bibitem[{OpenAI et~al.(2024)OpenAI, Achiam, Adler, Agarwal, Ahmad, Akkaya, Aleman, Almeida, Altenschmidt, Altman, Anadkat, Avila, Babuschkin, Balaji, Balcom, Baltescu, Bao, Bavarian, Belgum, Bello, Berdine, Bernadett-Shapiro, Berner, Bogdonoff, Boiko, Boyd, Brakman, Brockman, Brooks, Brundage, Button, Cai, Campbell, Cann, Carey, Carlson, Carmichael, Chan, Chang, Chantzis, Chen, Chen, Chen, Chen, Chen, Chess, Cho, Chu, Chung, Cummings, Currier, Dai, Decareaux, Degry, Deutsch, Deville, Dhar, Dohan, Dowling, Dunning, Ecoffet, Eleti, Eloundou, Farhi, Fedus, Felix, Fishman, Forte, Fulford, Gao, Georges, Gibson, Goel, Gogineni, Goh, Gontijo-Lopes, Gordon, Grafstein, Gray, Greene, Gross, Gu, Guo, Hallacy, Han, Harris, He, Heaton, Heidecke, Hesse, Hickey, Hickey, Hoeschele, Houghton, Hsu, Hu, Hu, Huizinga, Jain, Jain, Jang, Jiang, Jiang, Jin, Jin, Jomoto, Jonn, Jun, Kaftan, Łukasz Kaiser, Kamali, Kanitscheider, Keskar, Khan, Kilpatrick, Kim, Kim, Kim, Kirchner, Kiros, Knight, Kokotajlo, Łukasz Kondraciuk,
  Kondrich, Konstantinidis, Kosic, Krueger, Kuo, Lampe, Lan, Lee, Leike, Leung, Levy, Li, Lim, Lin, Lin, Litwin, Lopez, Lowe, Lue, Makanju, Malfacini, Manning, Markov, Markovski, Martin, Mayer, Mayne, McGrew, McKinney, McLeavey, McMillan, McNeil, Medina, Mehta, Menick, Metz, Mishchenko, Mishkin, Monaco, Morikawa, Mossing, Mu, Murati, Murk, Mély, Nair, Nakano, Nayak, Neelakantan, Ngo, Noh, Ouyang, O'Keefe, Pachocki, Paino, Palermo, Pantuliano, Parascandolo, Parish, Parparita, Passos, Pavlov, Peng, Perelman, de~Avila Belbute~Peres, Petrov, de~Oliveira~Pinto, Michael, Pokorny, Pokrass, Pong, Powell, Power, Power, Proehl, Puri, Radford, Rae, Ramesh, Raymond, Real, Rimbach, Ross, Rotsted, Roussez, Ryder, Saltarelli, Sanders, Santurkar, Sastry, Schmidt, Schnurr, Schulman, Selsam, Sheppard, Sherbakov, Shieh, Shoker, Shyam, Sidor, Sigler, Simens, Sitkin, Slama, Sohl, Sokolowsky, Song, Staudacher, Such, Summers, Sutskever, Tang, Tezak, Thompson, Tillet, Tootoonchian, Tseng, Tuggle, Turley, Tworek, Uribe, Vallone,
  Vijayvergiya, Voss, Wainwright, Wang, Wang, Wang, Ward, Wei, Weinmann, Welihinda, Welinder, Weng, Weng, Wiethoff, Willner, Winter, Wolrich, Wong, Workman, Wu, Wu, Wu, Xiao, Xu, Yoo, Yu, Yuan, Zaremba, Zellers, Zhang, Zhang, Zhao, Zheng, Zhuang, Zhuk, and Zoph}]{openai2024gpt4}
OpenAI, Josh Achiam, Steven Adler, Sandhini Agarwal, Lama Ahmad, Ilge Akkaya, Florencia~Leoni Aleman, Diogo Almeida, Janko Altenschmidt, Sam Altman, Shyamal Anadkat, Red Avila, Igor Babuschkin, Suchir Balaji, Valerie Balcom, Paul Baltescu, Haiming Bao, Mohammad Bavarian, Jeff Belgum, Irwan Bello, Jake Berdine, Gabriel Bernadett-Shapiro, Christopher Berner, Lenny Bogdonoff, Oleg Boiko, Madelaine Boyd, Anna-Luisa Brakman, Greg Brockman, Tim Brooks, Miles Brundage, Kevin Button, Trevor Cai, Rosie Campbell, Andrew Cann, Brittany Carey, Chelsea Carlson, Rory Carmichael, Brooke Chan, Che Chang, Fotis Chantzis, Derek Chen, Sully Chen, Ruby Chen, Jason Chen, Mark Chen, Ben Chess, Chester Cho, Casey Chu, Hyung~Won Chung, Dave Cummings, Jeremiah Currier, Yunxing Dai, Cory Decareaux, Thomas Degry, Noah Deutsch, Damien Deville, Arka Dhar, David Dohan, Steve Dowling, Sheila Dunning, Adrien Ecoffet, Atty Eleti, Tyna Eloundou, David Farhi, Liam Fedus, Niko Felix, Simón~Posada Fishman, Juston Forte, Isabella Fulford, Leo
  Gao, Elie Georges, Christian Gibson, Vik Goel, Tarun Gogineni, Gabriel Goh, Rapha Gontijo-Lopes, Jonathan Gordon, Morgan Grafstein, Scott Gray, Ryan Greene, Joshua Gross, Shixiang~Shane Gu, Yufei Guo, Chris Hallacy, Jesse Han, Jeff Harris, Yuchen He, Mike Heaton, Johannes Heidecke, Chris Hesse, Alan Hickey, Wade Hickey, Peter Hoeschele, Brandon Houghton, Kenny Hsu, Shengli Hu, Xin Hu, Joost Huizinga, Shantanu Jain, Shawn Jain, Joanne Jang, Angela Jiang, Roger Jiang, Haozhun Jin, Denny Jin, Shino Jomoto, Billie Jonn, Heewoo Jun, Tomer Kaftan, Łukasz Kaiser, Ali Kamali, Ingmar Kanitscheider, Nitish~Shirish Keskar, Tabarak Khan, Logan Kilpatrick, Jong~Wook Kim, Christina Kim, Yongjik Kim, Jan~Hendrik Kirchner, Jamie Kiros, Matt Knight, Daniel Kokotajlo, Łukasz Kondraciuk, Andrew Kondrich, Aris Konstantinidis, Kyle Kosic, Gretchen Krueger, Vishal Kuo, Michael Lampe, Ikai Lan, Teddy Lee, Jan Leike, Jade Leung, Daniel Levy, Chak~Ming Li, Rachel Lim, Molly Lin, Stephanie Lin, Mateusz Litwin, Theresa Lopez, Ryan
  Lowe, Patricia Lue, Anna Makanju, Kim Malfacini, Sam Manning, Todor Markov, Yaniv Markovski, Bianca Martin, Katie Mayer, Andrew Mayne, Bob McGrew, Scott~Mayer McKinney, Christine McLeavey, Paul McMillan, Jake McNeil, David Medina, Aalok Mehta, Jacob Menick, Luke Metz, Andrey Mishchenko, Pamela Mishkin, Vinnie Monaco, Evan Morikawa, Daniel Mossing, Tong Mu, Mira Murati, Oleg Murk, David Mély, Ashvin Nair, Reiichiro Nakano, Rajeev Nayak, Arvind Neelakantan, Richard Ngo, Hyeonwoo Noh, Long Ouyang, Cullen O'Keefe, Jakub Pachocki, Alex Paino, Joe Palermo, Ashley Pantuliano, Giambattista Parascandolo, Joel Parish, Emy Parparita, Alex Passos, Mikhail Pavlov, Andrew Peng, Adam Perelman, Filipe de~Avila Belbute~Peres, Michael Petrov, Henrique~Ponde de~Oliveira~Pinto, Michael, Pokorny, Michelle Pokrass, Vitchyr~H. Pong, Tolly Powell, Alethea Power, Boris Power, Elizabeth Proehl, Raul Puri, Alec Radford, Jack Rae, Aditya Ramesh, Cameron Raymond, Francis Real, Kendra Rimbach, Carl Ross, Bob Rotsted, Henri Roussez,
  Nick Ryder, Mario Saltarelli, Ted Sanders, Shibani Santurkar, Girish Sastry, Heather Schmidt, David Schnurr, John Schulman, Daniel Selsam, Kyla Sheppard, Toki Sherbakov, Jessica Shieh, Sarah Shoker, Pranav Shyam, Szymon Sidor, Eric Sigler, Maddie Simens, Jordan Sitkin, Katarina Slama, Ian Sohl, Benjamin Sokolowsky, Yang Song, Natalie Staudacher, Felipe~Petroski Such, Natalie Summers, Ilya Sutskever, Jie Tang, Nikolas Tezak, Madeleine~B. Thompson, Phil Tillet, Amin Tootoonchian, Elizabeth Tseng, Preston Tuggle, Nick Turley, Jerry Tworek, Juan Felipe~Cerón Uribe, Andrea Vallone, Arun Vijayvergiya, Chelsea Voss, Carroll Wainwright, Justin~Jay Wang, Alvin Wang, Ben Wang, Jonathan Ward, Jason Wei, CJ~Weinmann, Akila Welihinda, Peter Welinder, Jiayi Weng, Lilian Weng, Matt Wiethoff, Dave Willner, Clemens Winter, Samuel Wolrich, Hannah Wong, Lauren Workman, Sherwin Wu, Jeff Wu, Michael Wu, Kai Xiao, Tao Xu, Sarah Yoo, Kevin Yu, Qiming Yuan, Wojciech Zaremba, Rowan Zellers, Chong Zhang, Marvin Zhang, Shengjia
  Zhao, Tianhao Zheng, Juntang Zhuang, William Zhuk, and Barret Zoph. 2024.
\newblock \href {https://arxiv.org/abs/2303.08774} {Gpt-4 technical report}.
\newblock \emph{Preprint}, arXiv:2303.08774.

\bibitem[{Panickssery et~al.(2024)Panickssery, Bowman, and Feng}]{panickssery2024llm}
Arjun Panickssery, Samuel~R. Bowman, and Shi Feng. 2024.
\newblock \href {https://arxiv.org/abs/2404.13076} {Llm evaluators recognize and favor their own generations}.
\newblock \emph{Preprint}, arXiv:2404.13076.

\bibitem[{Radev et~al.(2002)Radev, Qi, Wu, and Fan}]{radev-etal-2002-evaluating}
Dragomir~R. Radev, Hong Qi, Harris Wu, and Weiguo Fan. 2002.
\newblock \href {http://www.lrec-conf.org/proceedings/lrec2002/pdf/301.pdf} {Evaluating web-based question answering systems}.
\newblock In \emph{Proceedings of the Third International Conference on Language Resources and Evaluation ({LREC}{'}02)}, Las Palmas, Canary Islands - Spain. European Language Resources Association (ELRA).

\bibitem[{Raffel et~al.(2020)Raffel, Shazeer, Roberts, Lee, Narang, Matena, Zhou, Li, and Liu}]{T5}
Colin Raffel, Noam Shazeer, Adam Roberts, Katherine Lee, Sharan Narang, Michael Matena, Yanqi Zhou, Wei Li, and Peter~J. Liu. 2020.
\newblock Exploring the limits of transfer learning with a unified text-to-text transformer.
\newblock \emph{J. Mach. Learn. Res.}, 21(1).

\bibitem[{Ranaldi and Freitas(2024)}]{ranaldi-freitas-2024-aligning}
Leonardo Ranaldi and Andre Freitas. 2024.
\newblock \href {https://aclanthology.org/2024.eacl-long.109} {Aligning large and small language models via chain-of-thought reasoning}.
\newblock In \emph{Proceedings of the 18th Conference of the European Chapter of the Association for Computational Linguistics (Volume 1: Long Papers)}, pages 1812--1827, St. Julian{'}s, Malta. Association for Computational Linguistics.

\bibitem[{Reimers and Gurevych(2019)}]{reimers-gurevych-2019-sentence}
Nils Reimers and Iryna Gurevych. 2019.
\newblock \href {https://doi.org/10.18653/v1/D19-1410} {Sentence-{BERT}: Sentence embeddings using {S}iamese {BERT}-networks}.
\newblock In \emph{Proceedings of the 2019 Conference on Empirical Methods in Natural Language Processing and the 9th International Joint Conference on Natural Language Processing (EMNLP-IJCNLP)}, pages 3982--3992, Hong Kong, China. Association for Computational Linguistics.

\bibitem[{Shen et~al.(2023)Shen, Cheng, Nguyen, You, and Bing}]{shen-etal-2023-large}
Chenhui Shen, Liying Cheng, Xuan-Phi Nguyen, Yang You, and Lidong Bing. 2023.
\newblock \href {https://doi.org/10.18653/v1/2023.findings-emnlp.278} {Large language models are not yet human-level evaluators for abstractive summarization}.
\newblock In \emph{Findings of the Association for Computational Linguistics: EMNLP 2023}, pages 4215--4233, Singapore. Association for Computational Linguistics.

\bibitem[{Soleimani et~al.(2022)Soleimani, Nikoulina, Favre, and Ait~Mokhtar}]{soleimani-etal-2022-zero}
Amir Soleimani, Vassilina Nikoulina, Benoit Favre, and Salah Ait~Mokhtar. 2022.
\newblock \href {https://doi.org/10.18653/v1/2022.bionlp-1.5} {Zero-shot aspect-based scientific document summarization using self-supervised pre-training}.
\newblock In \emph{Proceedings of the 21st Workshop on Biomedical Language Processing}, pages 49--62, Dublin, Ireland. Association for Computational Linguistics.

\bibitem[{Song et~al.(2020)Song, Tan, Qin, Lu, and Liu}]{10.5555/3495724.3497138}
Kaitao Song, Xu~Tan, Tao Qin, Jianfeng Lu, and Tie-Yan Liu. 2020.
\newblock Mpnet: masked and permuted pre-training for language understanding.
\newblock In \emph{Proceedings of the 34th International Conference on Neural Information Processing Systems}, NIPS '20, Red Hook, NY, USA. Curran Associates Inc.

\bibitem[{Stede and Patz(2021)}]{stede-patz-2021-climate}
Manfred Stede and Ronny Patz. 2021.
\newblock \href {https://doi.org/10.18653/v1/2021.nlp4posimpact-1.2} {The climate change debate and natural language processing}.
\newblock In \emph{Proceedings of the 1st Workshop on NLP for Positive Impact}, pages 8--18, Online. Association for Computational Linguistics.

\bibitem[{Tan et~al.(2020)Tan, Qin, Xing, and Hu}]{tan-etal-2020-summarizing}
Bowen Tan, Lianhui Qin, Eric Xing, and Zhiting Hu. 2020.
\newblock \href {https://doi.org/10.18653/v1/2020.emnlp-main.510} {Summarizing text on any aspects: A knowledge-informed weakly-supervised approach}.
\newblock In \emph{Proceedings of the 2020 Conference on Empirical Methods in Natural Language Processing (EMNLP)}, pages 6301--6309, Online. Association for Computational Linguistics.

\bibitem[{Team et~al.(2024)Team, Mesnard, Hardin, Dadashi, Bhupatiraju, Pathak, Sifre, Rivière, Kale, Love, Tafti, Hussenot, Sessa, Chowdhery, Roberts, Barua, Botev, Castro-Ros, Slone, Héliou, Tacchetti, Bulanova, Paterson, Tsai, Shahriari, Lan, Choquette-Choo, Crepy, Cer, Ippolito, Reid, Buchatskaya, Ni, Noland, Yan, Tucker, Muraru, Rozhdestvenskiy, Michalewski, Tenney, Grishchenko, Austin, Keeling, Labanowski, Lespiau, Stanway, Brennan, Chen, Ferret, Chiu, Mao-Jones, Lee, Yu, Millican, Sjoesund, Lee, Dixon, Reid, Mikuła, Wirth, Sharman, Chinaev, Thain, Bachem, Chang, Wahltinez, Bailey, Michel, Yotov, Chaabouni, Comanescu, Jana, Anil, McIlroy, Liu, Mullins, Smith, Borgeaud, Girgin, Douglas, Pandya, Shakeri, De, Klimenko, Hennigan, Feinberg, Stokowiec, hui Chen, Ahmed, Gong, Warkentin, Peran, Giang, Farabet, Vinyals, Dean, Kavukcuoglu, Hassabis, Ghahramani, Eck, Barral, Pereira, Collins, Joulin, Fiedel, Senter, Andreev, and Kenealy}]{gemmateam2024gemma}
Gemma Team, Thomas Mesnard, Cassidy Hardin, Robert Dadashi, Surya Bhupatiraju, Shreya Pathak, Laurent Sifre, Morgane Rivière, Mihir~Sanjay Kale, Juliette Love, Pouya Tafti, Léonard Hussenot, Pier~Giuseppe Sessa, Aakanksha Chowdhery, Adam Roberts, Aditya Barua, Alex Botev, Alex Castro-Ros, Ambrose Slone, Amélie Héliou, Andrea Tacchetti, Anna Bulanova, Antonia Paterson, Beth Tsai, Bobak Shahriari, Charline~Le Lan, Christopher~A. Choquette-Choo, Clément Crepy, Daniel Cer, Daphne Ippolito, David Reid, Elena Buchatskaya, Eric Ni, Eric Noland, Geng Yan, George Tucker, George-Christian Muraru, Grigory Rozhdestvenskiy, Henryk Michalewski, Ian Tenney, Ivan Grishchenko, Jacob Austin, James Keeling, Jane Labanowski, Jean-Baptiste Lespiau, Jeff Stanway, Jenny Brennan, Jeremy Chen, Johan Ferret, Justin Chiu, Justin Mao-Jones, Katherine Lee, Kathy Yu, Katie Millican, Lars~Lowe Sjoesund, Lisa Lee, Lucas Dixon, Machel Reid, Maciej Mikuła, Mateo Wirth, Michael Sharman, Nikolai Chinaev, Nithum Thain, Olivier Bachem,
  Oscar Chang, Oscar Wahltinez, Paige Bailey, Paul Michel, Petko Yotov, Rahma Chaabouni, Ramona Comanescu, Reena Jana, Rohan Anil, Ross McIlroy, Ruibo Liu, Ryan Mullins, Samuel~L Smith, Sebastian Borgeaud, Sertan Girgin, Sholto Douglas, Shree Pandya, Siamak Shakeri, Soham De, Ted Klimenko, Tom Hennigan, Vlad Feinberg, Wojciech Stokowiec, Yu~hui Chen, Zafarali Ahmed, Zhitao Gong, Tris Warkentin, Ludovic Peran, Minh Giang, Clément Farabet, Oriol Vinyals, Jeff Dean, Koray Kavukcuoglu, Demis Hassabis, Zoubin Ghahramani, Douglas Eck, Joelle Barral, Fernando Pereira, Eli Collins, Armand Joulin, Noah Fiedel, Evan Senter, Alek Andreev, and Kathleen Kenealy. 2024.
\newblock \href {https://arxiv.org/abs/2403.08295} {Gemma: Open models based on gemini research and technology}.
\newblock \emph{Preprint}, arXiv:2403.08295.

\bibitem[{Titov and McDonald(2008)}]{titov-mcdonald-2008-joint}
Ivan Titov and Ryan McDonald. 2008.
\newblock \href {https://aclanthology.org/P08-1036} {A joint model of text and aspect ratings for sentiment summarization}.
\newblock In \emph{Proceedings of ACL-08: HLT}, pages 308--316, Columbus, Ohio. Association for Computational Linguistics.

\bibitem[{Tokayev(2023)}]{Tokayev_2023}
Kassym-Jomart Tokayev. 2023.
\newblock \href {https://norislab.com/index.php/ijsa/article/view/42} {Ethical implications of large language models a multidimensional exploration of societal, economic, and technical concerns}.
\newblock \emph{International Journal of Social Analytics}, 8(9):17–33.

\bibitem[{Vaghefi et~al.(2023)Vaghefi, Stammbach, Muccione, Bingler, Ni, Kraus, Allen, Colesanti-Senni, Wekhof, Schimanski, Gostlow, Yu, Wang, Webersinke, Huggel, and Leippold}]{Vaghefi2023-fc}
Saeid~Ashraf Vaghefi, Dominik Stammbach, Veruska Muccione, Julia Bingler, Jingwei Ni, Mathias Kraus, Simon Allen, Chiara Colesanti-Senni, Tobias Wekhof, Tobias Schimanski, Glen Gostlow, Tingyu Yu, Qian Wang, Nicolas Webersinke, Christian Huggel, and Markus Leippold. 2023.
\newblock {ChatClimate}: Grounding conversational {AI} in climate science.
\newblock \emph{Commun. Earth Environ.}, 4(1).

\bibitem[{Wan et~al.(2024)Wan, Wang, Liu, Alam, Zheng, Liu, Qu, Yan, Zhu, Zhang, Chowdhury, and Zhang}]{wan2024efficient}
Zhongwei Wan, Xin Wang, Che Liu, Samiul Alam, Yu~Zheng, Jiachen Liu, Zhongnan Qu, Shen Yan, Yi~Zhu, Quanlu Zhang, Mosharaf Chowdhury, and Mi~Zhang. 2024.
\newblock \href {https://arxiv.org/abs/2312.03863} {Efficient large language models: A survey}.
\newblock \emph{Preprint}, arXiv:2312.03863.

\bibitem[{Yang et~al.(2023)Yang, Song, Cho, Wang, Pan, Petzold, and Yu}]{yang-etal-2023-oasum}
Xianjun Yang, Kaiqiang Song, Sangwoo Cho, Xiaoyang Wang, Xiaoman Pan, Linda Petzold, and Dong Yu. 2023.
\newblock \href {https://doi.org/10.18653/v1/2023.findings-acl.268} {{OAS}um: Large-scale open domain aspect-based summarization}.
\newblock In \emph{Findings of the Association for Computational Linguistics: ACL 2023}, pages 4381--4401, Toronto, Canada. Association for Computational Linguistics.

\bibitem[{Yao et~al.(2024)Yao, Wu, Li, Youn, and He}]{Yao_Wu_Li_Youn_He_2024}
Zhewei Yao, Xiaoxia Wu, Cheng Li, Stephen Youn, and Yuxiong He. 2024.
\newblock \href {https://doi.org/10.1609/aaai.v38i17.29908} {Exploring post-training quantization in llms from comprehensive study to low rank compensation}.
\newblock \emph{Proceedings of the AAAI Conference on Artificial Intelligence}, 38(17):19377--19385.

\bibitem[{Zhang et~al.(2024)Zhang, Ladhak, Durmus, Liang, McKeown, and Hashimoto}]{10.1162/tacl_a_00632}
Tianyi Zhang, Faisal Ladhak, Esin Durmus, Percy Liang, Kathleen McKeown, and Tatsunori~B. Hashimoto. 2024.
\newblock \href {https://doi.org/10.1162/tacl_a_00632} {{Benchmarking Large Language Models for News Summarization}}.
\newblock \emph{Transactions of the Association for Computational Linguistics}, 12:39--57.

\bibitem[{Ziyu et~al.(2023)Ziyu, Qiguang, Longxuan, Mingda, Yi, Yushan, Haopeng, Weinan, and Liu}]{ziyu-etal-2023-lens}
Zhuang Ziyu, Chen Qiguang, Ma~Longxuan, Li~Mingda, Han Yi, Qian Yushan, Bai Haopeng, Zhang Weinan, and Ting Liu. 2023.
\newblock \href {https://aclanthology.org/2023.ccl-2.8} {Through the lens of core competency: Survey on evaluation of large language models}.
\newblock In \emph{Proceedings of the 22nd Chinese National Conference on Computational Linguistics (Volume 2: Frontier Forum)}, pages 88--109, Harbin, China. Chinese Information Processing Society of China.

\end{thebibliography}

\appendix

\section{Metric Correlation with Human Judgement}
\label{sec:appendixA}

Previous research has variously shown how summarization metrics are generally unreliable, yielding low correlation with human judgement; the use of ChatGPT in this context was observed to be the method yielding results more similar to the judgement expressed by human annotators, with correlation  values around 0.50 \cite{shen-etal-2023-large}. Still, our use case was slightly different from the one in the above work, as it deals with ABS rather than normal summarization and, given the specificity of our dataset (see appendix \ref{sec:appendixA}) it also includes various snippets of texts directly copied from the main text in the reference summaries.

To assess the reliability of different metrics in this context and to choose which to report, we have asked two human annotators to rank 10 pairs of summaries generated by different LLMs and then we compared the results thus obtained with the ranking produced by different summarization metrics. Table \ref{tab:metricagr} shows the percentage of matches between human annotators' rankings and the metrics obtained by recent metrics based on LLMs. It can be seen how ChatGPT RTS far outperforms the alternatives reaching very high agreement with the human annotators (close to 80\%).

\begin{table*}
\centering
\small
    \begin{tabular}{c|c|c|c|c}
         Metric & Consistency & Coherence & Fluency & Relevance \\
         ChatGPT RTS & $0.77\pm 0.0$ & $0.83 \pm 0.06$ & $0.66 \pm 0.11$ & $0.77 \pm 0.0$ \\ 
         ChatGPT MCQ & $0.06 \pm 0.06$ & $0.55 \pm 0.0$ & $0.17 \pm 0.06$ & $0.44 \pm 0.0$ \\
         UniEval& $55 \pm 0.11$ & $0.61 \pm 0.06$ & $0.33 \pm 0.22$ & $0.67 \pm 0.0$\\
    \end{tabular}
    \caption{Average percentage of agreement between human annotators and LLM-based summarization metrics: standard deviation is also included.}
    \label{tab:metricagr}
\end{table*}

\begin{figure}
    \centering
    \includegraphics[width=18em, height=18em]{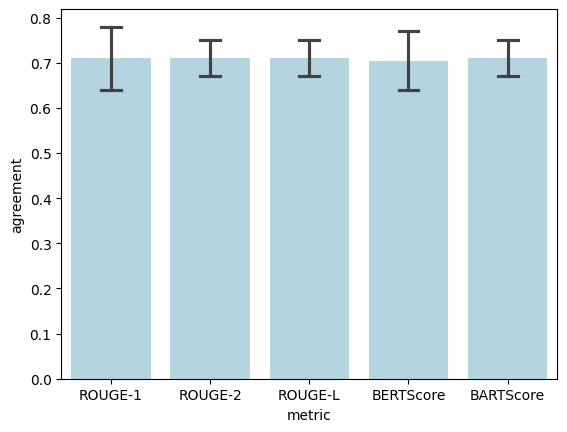}
    \caption{Average percentage of agreement between human annotators and similarity-based summarization metrics: standard deviation is also included in the form of error bars.}
    \label{fig:metricagr}
\end{figure}

If we consider the agreement with traditional, similarity-based metrics depicted in figure \ref{fig:metricagr}, we can also observe how the the majority of traditional metrics generally agree with human annotators in this task at a level close to the one reached by ChatGPT RTS. This is indeed quite specific to the dataset we are considering as summaries are often presented as highlights reporting entire sentences from the source paragraph and, as LLMs are asked to generate highlights as well, rather than summaries, similarity-based metrics are actually quite good in this scenario. As traditional metrics lack a distinction between different dimensions of the generated summaries, however, we opted for ChatGPT RTS as the metric for our main experiments. 


\section{Evaluation Prompts}
\label{sec:appendixB}
In using the ChatGPT RTS, we have prompted ChatGPT with 4 different prompts per summary, to evaluate the different dimensions of the generated summaries. For what concerns consistency, coherence and fluency, we have adopted the same prompts from \citet{shen-etal-2023-large}. For what concerns relevance, we re-adapted the original formulation to make it fit for ABS, where we want our summary to be relevant with respect to a specific topic, in addition to the reference summary, where the original formulation did not include any topic nor reference summary.

\begin{figure}
    \includegraphics[width=20em, height=40em]{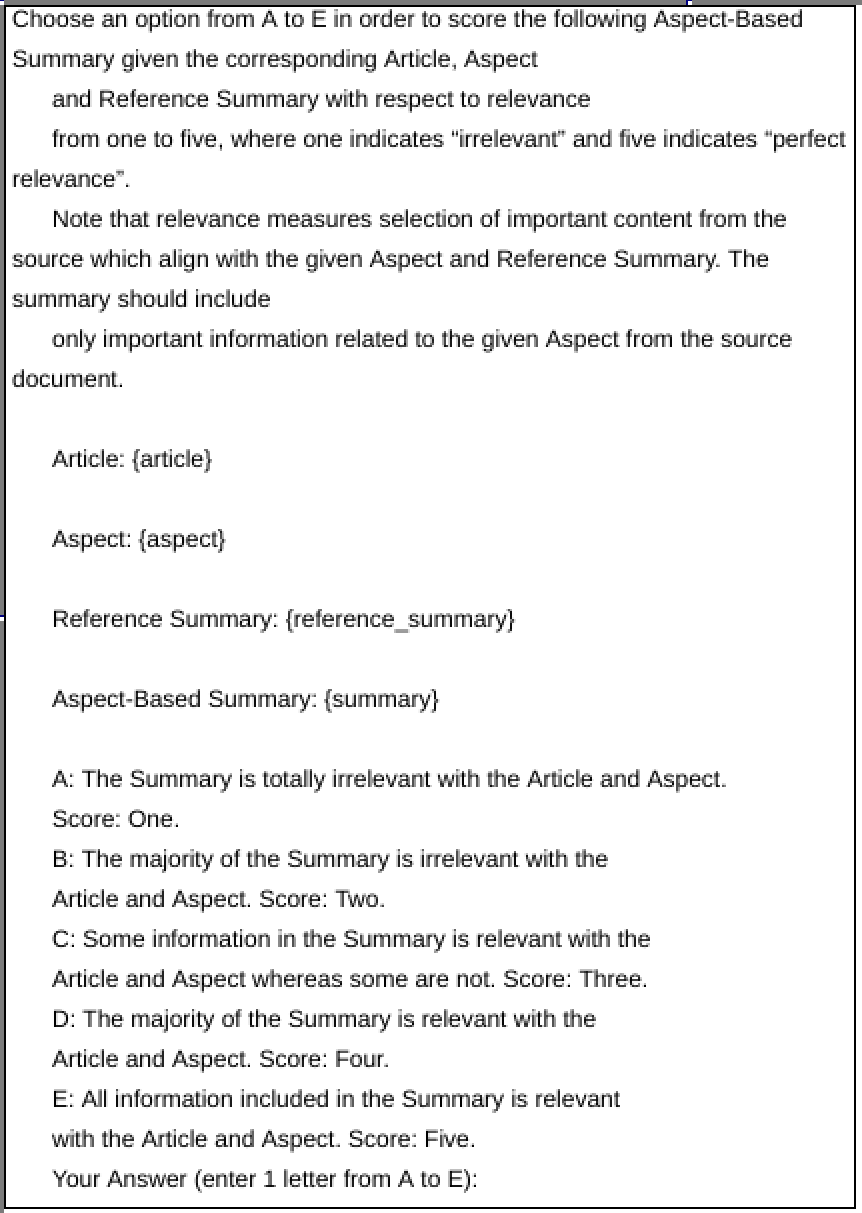}
    \caption{The prompt used for evaluation with ChatGPT with the ChatGPT RTS evaluation method for the relevance aspect. At inference time \{article\} is substituted with the target paragraphs, \{aspect\} is substituted with the aspect on which the summarizer should focus, \{reference\_summary\} is replaced with the reference summary and \{summary\} is replaced with the generated summary. All other dimensions have been evaluated with similar prompts, but without the need of \{reference\_summary\} and \{aspect\} and substituting the description of the dimension with the relevant description from the other dimensions, as described in \citet{shen-etal-2023-large}.}
    \label{fig:promptrelevance}
\end{figure}

We refer the reader to the original formulation in \citet{shen-etal-2023-large} for the prompt used for consistency, coherence and fluency dimensions. For the relevance dimension, we show the prompt we used in figure \ref{fig:promptrelevance}.

\section{Effect of Long Inputs}
\label{sec:appendixC}
In the methodology section, we highlighted how when using SLMs for summarization is usual to find instances in which input paragraphs are longer than the allowed token limits for the model. We have tackled these instances by applying an iterative procedure where we summarize individual paragraphs and then we ask the given LLM to summarize the concatenation of the summaries. 

Formally, we set a character threshold over which we get a set of interim results $y_{int}^p$:
\begin{equation}
    y_{int}^p=LLM(sub(T, topic, p)) \forall p \in P
\label{eq:long}
\end{equation}

Then, having the collection $Y_{int}$ of all $y_{int}^p$, we get the final text as:
\begin{equation}
    text=concat(Y_{int})
    \label{eq:long2}
\end{equation}

which can then be passed in equation \ref{eq:prompt} to obtain the final prompt to be passed in equation \ref{eq:gen}.

\begin{figure}
    \centering
    \includegraphics[width=20em, height=17em]{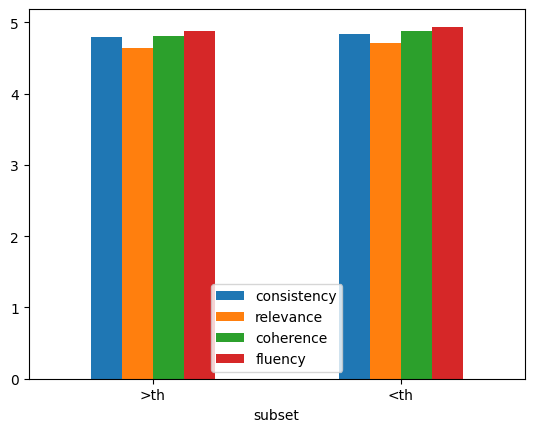}
    \caption{Average comparison of performance across models in terms of ChatGPT RTS for instances longer (left) and shorter (right) than our fixed threshold (th).}
    \label{fig:longshort}
\end{figure}

In order to ensure that such a process won't lead to drop in performance we have plotted the performance of instances in which the paragraphs are longer than the fixed character threshold that we have empirically set in our experiments and we compared them to the plotted performance of instances where paragraphs are shorter than the threshold. Figure \ref{fig:longshort} shows such comparison: as it can be noticed, the difference in performance shows longer inputs performing marginally worse due to the iterative procedure. Still, the difference is minimal suggesting that our method does not negatively affect performance and, as such, it is a feasible way to overcome models' context limits.

\section{Qualitative Evaluation}
\label{sec:appendixD}

\begin{figure}[!thb]
    \includegraphics[scale=0.18]{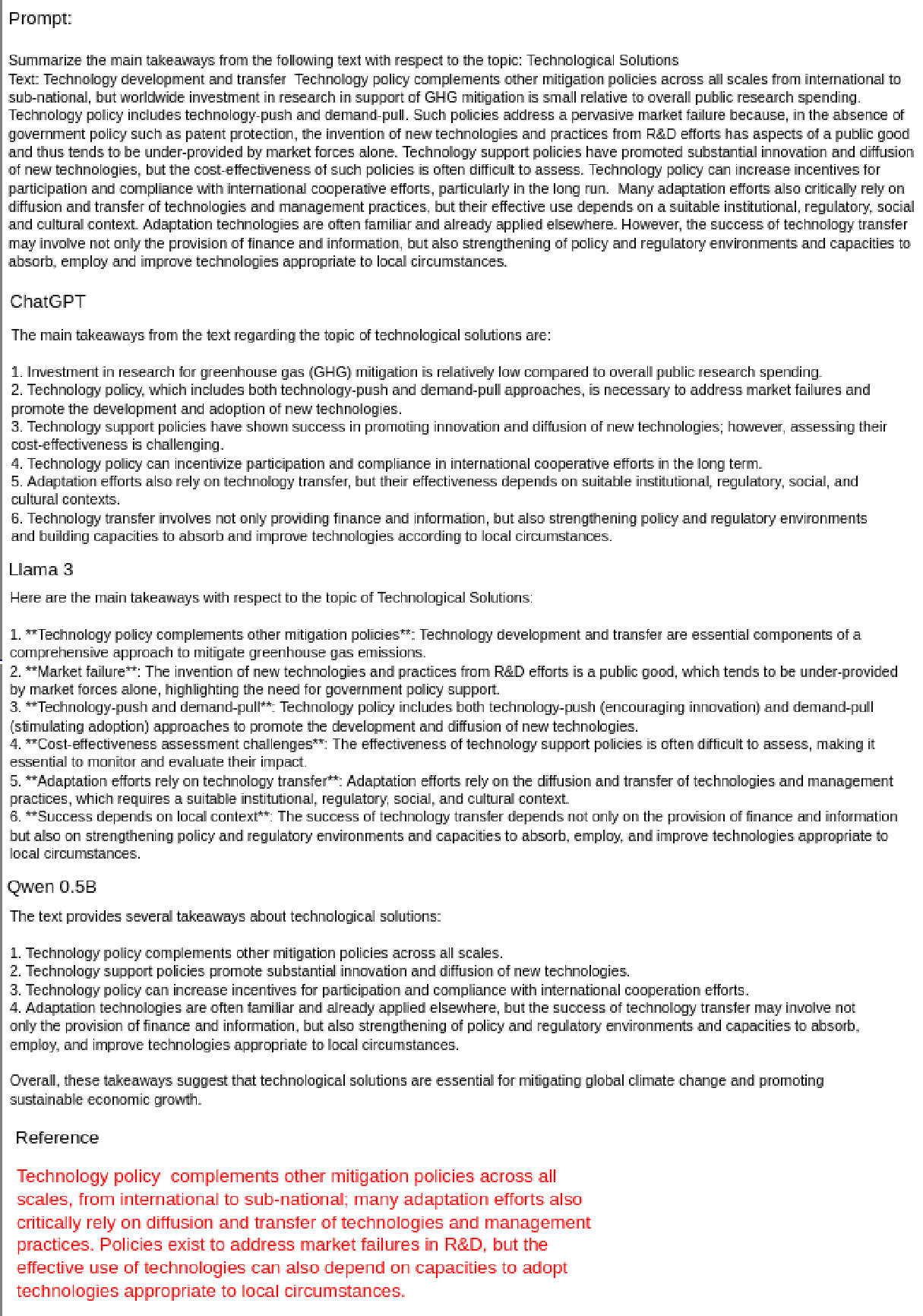}
    \caption{An example of three summaries obtained using three LLMs: ChatGPT, Llama 3 and Qwen 0.5B. Prompt indicates the command given to the LLMs, including the text to be summarized and the target aspect. Reference indicates the reference human-generated summary. It can be seen how all models, even the smaller Qwen 0.5B, manage to produce sensible summaries, even though they do include extra information with respect to the reference summary (for which a more specific aspect formulation might be needed).}
    \label{fig:qualitative-example}
\end{figure}

\begin{figure}[!tbh]
    \includegraphics[trim={0 0 0 0.5cm},clip,width=25em, height=50em]{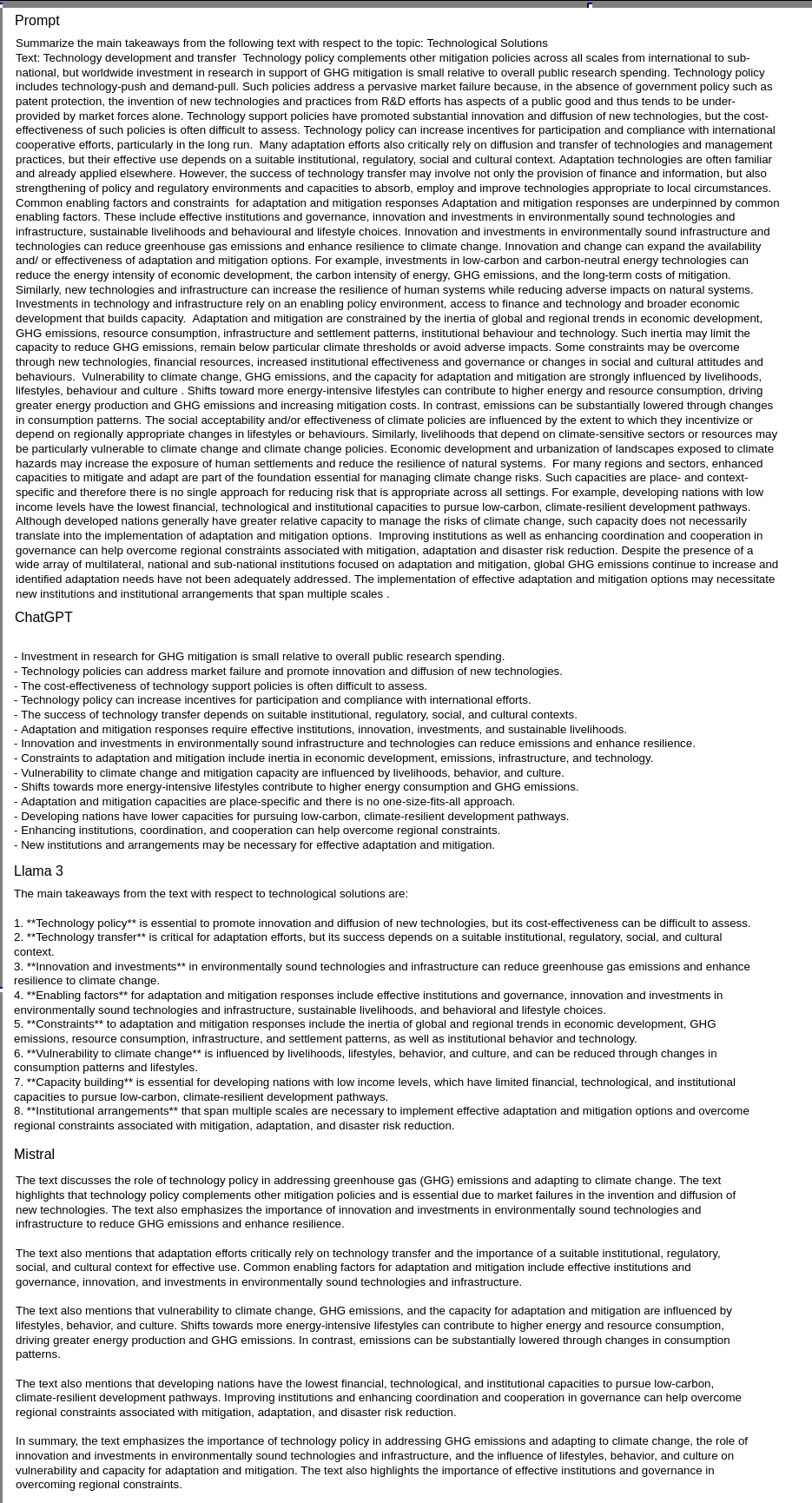}
    \caption{An example of three summaries obtained using three LLMs: ChatGPT, Llama 3 and Mistral (the weakest LLM among this set of experiments). Prompt indicates the command given to the LLMs, including the text to be summarized and the target aspect. The reference summary is depicted in figure \ref{fig:qualitative-example}.}
    \label{fig:qualitative-example_rag}
\end{figure}

Figure \ref{fig:qualitative-example} show an example of summaries generated for a given reference by different LLMs, together with the reference summary and the prompt used to obtain the summaries, including the target ground truth paragraph to be summarized. When ground truth target paragraphs are included, it can be seen that all LLMs give sensible answers which are comparable to each other. Some redundant information is included in all cases, but specifying the aspect more strictly is likely to solve that problem. When retrieved paragraphs are used in the RAG setting, instead, LLMs struggle to produce sensible results, as the discrepancy between the input (incorrect) paragraph and the aspect to be summarized tend to confuse the models, as highlighted in figure \ref{fig:qualitative-example_rag}: this effect is stronger for weaker models as evident from the significance of the results in table \ref{tab:4} and, looking picture \ref{fig:qualitative-example_rag}, from the case of Mistral, which produced a summary which is relatively long and mostly unrelated to the target aspect.

\section{Dataset Statistics}
\label{sec:appendixE}
\begin{figure}
    \centering
    \includegraphics[width=15em, height=15em]{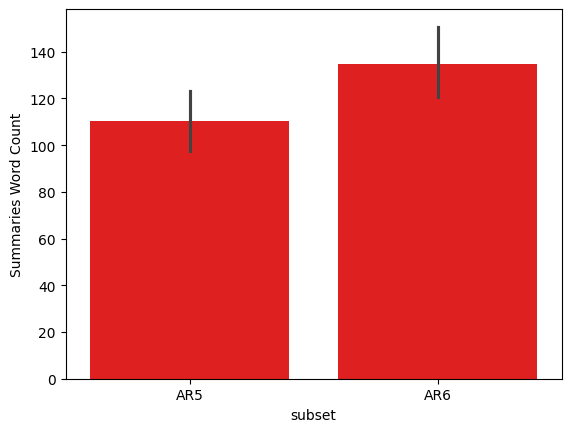}
    \caption{Average word count in the reference summaries for the two subsets of our dataset.}
    \label{fig:wordcount}
\end{figure}

\begin{figure}
    \centering
    \includegraphics[width=15em, height=15em]{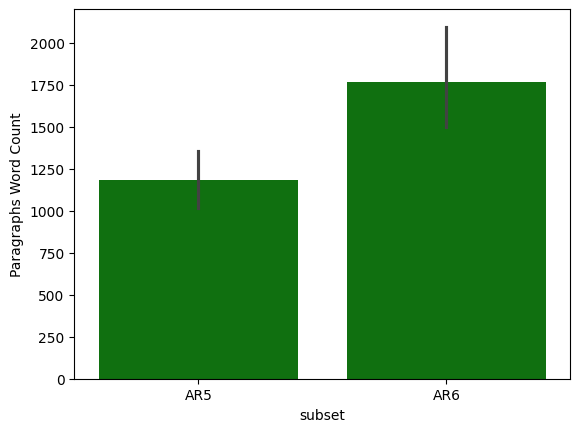}
    \caption{Average word count in the target paragraphs for the two subsets of our dataset.}
    \label{fig:wordparacount}
\end{figure}

\begin{figure}
    \centering
    \includegraphics[width=15em, height=15em]{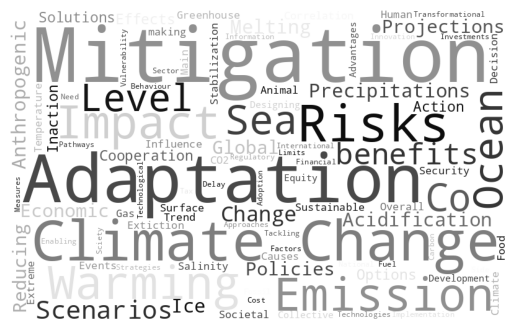}
    \caption{Most common summary topics in the AR5 subset of our dataset.}
    \label{fig:topicAR5}
\end{figure}

\begin{figure}
    \centering
    \includegraphics[width=15em, height=15em]{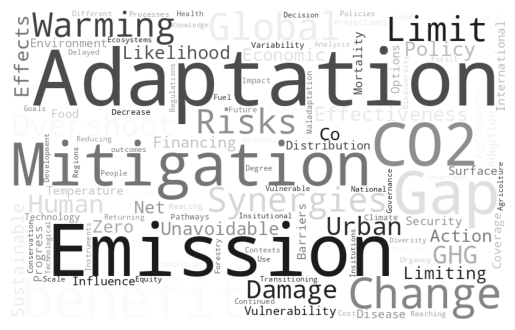}
    \caption{Most common summary topics in the AR6 subset of our dataset.}
    \label{fig:topicAR6}
\end{figure}

\begin{figure}
    \centering
    \includegraphics[width=18em, height=18em]{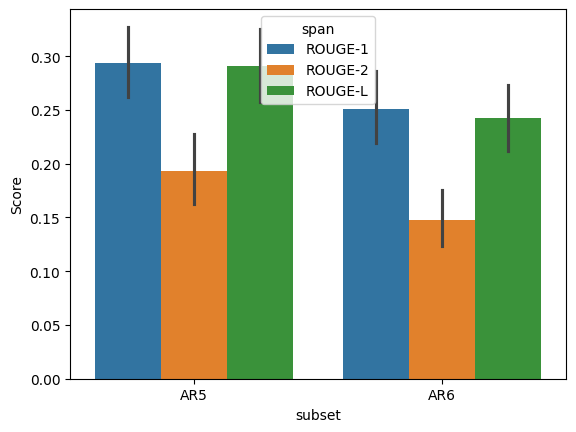}
    \caption{Rouge-1, rouge-2 and rouge-l scores of the reference summaries with respect to the target full paragraphs. These metrics represent the general overlap of the summaries with respect to the paragraphs, which is overall quite high in our case.}
    \label{fig:overlap}
\end{figure}

Here, we present more in depth statistics for our SumIPCC dataset which we release under MIT license.
Specifically, we report average word counts in summaries (figure \ref{fig:wordcount}) and in target paragraphs (figure \ref{fig:wordparacount}), more common words in the summaries' topics for AR5 (figure \ref{fig:topicAR5}) and AR6 (figure \ref{fig:topicAR6}) subsets and lexical overlaps between reference summaries and target paragraphs in terms of rouge-1, rouge-2 and rouge-l (figure \ref{fig:overlap}).

Overall, topics are similar between the two subsets and AR5 generally includes shorter paragraphs and shorter summaries than AR6. Also, it is evident by comparing figures \ref{fig:wordcount} and \ref{fig:wordparacount} how the compression rate is quite high. Finally, figure \ref{fig:overlap} show how the lexical overlap between reference summaries and target paragraphs is also quite high reflecting the nature of the summaries often reflecting highlights rather than abstractive summaries.

\section{Model Details}
\label{sec:appendixF}

In our experiments we have used in all cases the pre-trained models as hosted on Huggingface Hub, but for ChatGPT and GPT4, for which we have used the official API.

Specifically, we report below the link for each of the open-source models we used:

\begin{enumerate}
    \item Qwen 0.5B: \url{https://huggingface.co/Qwen/Qwen1.5-0.5B-Chat}
    \item Qwen 1.8B: \url{https://huggingface.co/Qwen/Qwen1.5-1.8B-Chat}
    \item Qwen 4B: \url{https://huggingface.co/Qwen/Qwen1.5-4B-Chat}
    \item Qwen 7B: \url{https://huggingface.co/Qwen/Qwen1.5-7B-Chat}
    \item Llama 3: \url{https://huggingface.co/meta-llama/Meta-Llama-3-8B}
    \item Gemma 2B: \url{https://huggingface.co/google/gemma-1.1-2b-it}
    \item Gemma 7B: \url{https://huggingface.co/google/gemma-1.1-7b-it}
    \item Phi 3: \url{https://huggingface.co/microsoft/Phi-3-mini-128k-instruct}
    \item Mistral: \url{https://huggingface.co/mistralai/Mistral-7B-Instruct-v0.2}
\end{enumerate}

The models were all quantized in 4 bit with the bitandbytes python library\footnote{https://github.com/TimDettmers/bitsandbytes} and run on a single NVIDIA® T4 GPU\footnote{https://www.nvidia.com/en-us/data-center/tesla-t4/} with 16GB of RAM, as previously explained. All the models run between 2.5 and 10 hours, depending on model size and length of generated summaries: no sampling was applied for replicability.

Details of the GPT models we used are presented in table \ref{tab:gpt}:
\begin{table}[!thb]
    \centering
    \small
    \begin{tabular}{c|c|c}
         Model& Model Official Name & Revision \\
         \hline
         ChatGPT& gpt-35-turbo-16k & 0613\\
         GPT4 & gpt-4 & 0125-Preview
    \end{tabular}
    \caption{Details of the used GPT models.}
    \label{tab:gpt}
\end{table}

Notice that throughout this work we have used the term ChatGPT to refer to GPT 3.5, consistently with previous literature \cite{shen-etal-2023-large}: this naming is, however, erroneous as ChatGPT refers to the service rather than the underlying model.

\end{document}